\documentclass[10pt,twocolumn,letterpaper]{article}

\usepackage[pagenumbers]{cvpr} 

\usepackage[dvipsnames]{xcolor}

\definecolor{cvprblue}{rgb}{0.21,0.49,0.74}
\usepackage[pagebackref,breaklinks,colorlinks,citecolor=cvprblue]{hyperref}
\usepackage{float}
\usepackage{wrapfig}
\usepackage{multirow}
\usepackage{color, colortbl}
\usepackage{graphicx}
\usepackage{soul}
\usepackage{marvosym}
\def\paperID{***} 
\def\confName{CVPR}
\def\confYear{2024}

\title{Driving into the Future: Multiview Visual Forecasting and Planning \\ with World Model for Autonomous Driving}
\author{
  Yuqi Wang$^{1*}$\quad 
  Jiawei He$^{1*}$ \quad
  Lue Fan$^{1*}$ \quad 
  Hongxin Li$^{1*}$ \quad
  Yuntao Chen$^{2}\textsuperscript{\Letter}$\quad 
  Zhaoxiang Zhang$^{1,2}\textsuperscript{\Letter}$ \quad
  \\ 
$^1$CASIA \quad $^2$CAIR, HKISI, CAS \\
  \small{Project Page: \url{https://drive-wm.github.io}} \\
  \small{Code: \url{https://github.com/BraveGroup/Drive-WM}}
}

\newcommand{\methodname}{Drive-WM}

\usepackage{amsmath}

\newcommand{\bc}{\mathbf{c}}
\newcommand{\x}{\mathbf{x}}
\newcommand{\y}{\mathbf{y}}
\newcommand{\f}{\mathbf{f}}

\newcommand{\I}{\boldsymbol{I}}
\newcommand{\0}{\mathbf{0}}

\newcommand{\beps}{\boldsymbol{\epsilon}}

\newcommand{\cN}{\mathcal{N}}

\newcommand{\bE}{\mathbb{E}}

\begin{document}

\twocolumn[{%
\vspace{-2em}
\maketitle%

{
    \centering
    \includegraphics[width=1\textwidth]{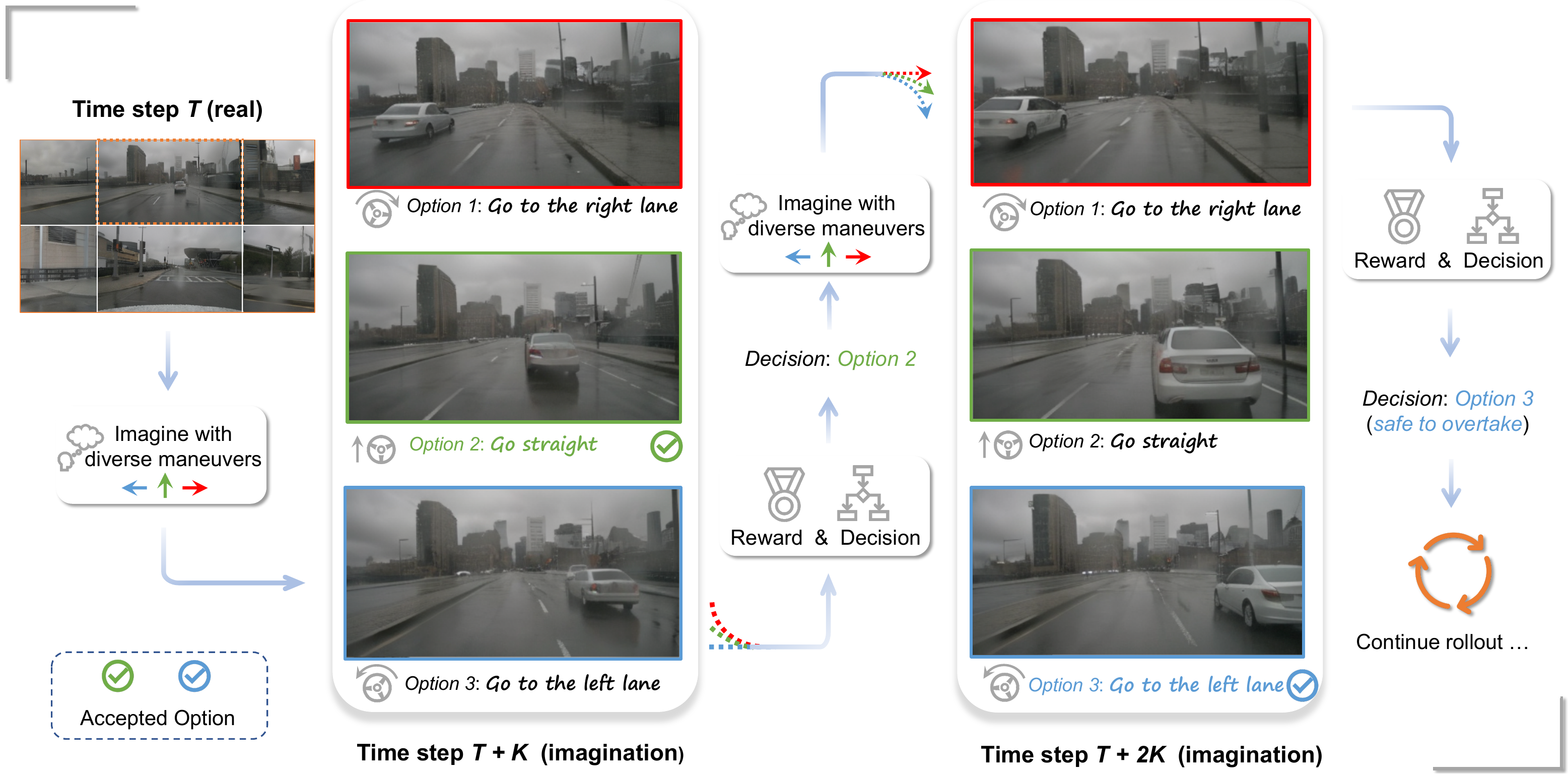}
    \captionof{figure}{\textbf{Multiview visual forecasting and planning by world model.}
    At time step $T$, the world model imagines the multiple futures at $T+K$, and finds it is safe to keep going straight at $T$. Then the model realizes that the ego car will be too close to the front car according to the imagination of time step $T + 2K$, so it decides to change to the left lane for a safe overtaking.}
    \label{fig:teaser}
    \vspace{1.5em}
}%
}]
{\let\thefootnote\relax\footnote{$^*$Equal contribution.}}
\begin{abstract}
In autonomous driving, predicting future events in advance and evaluating the foreseeable risks empowers autonomous vehicles to better plan their actions, enhancing safety and efficiency on the road.
To this end, we propose \textbf{\methodname{}}, the first driving world model compatible with existing end-to-end planning models.
Through a joint spatial-temporal modeling facilitated by view factorization, our model generates high-fidelity multiview videos in driving scenes.
Building on its powerful generation ability, we showcase the potential of applying the world model for safe driving planning for the first time. 
Particularly, our \textbf{\methodname{}} enables driving into multiple futures based on distinct driving maneuvers, and determines the optimal trajectory according to the image-based rewards.
Evaluation on real-world driving datasets verifies that our method could generate high-quality, consistent, and controllable multiview videos, opening up possibilities for real-world simulations and safe planning.
\end{abstract}

\section{Introduction}
\label{sec:introduction}

\begin{figure}[t]
    \centering
    \begin{subfigure}{0.48\columnwidth}
    \includegraphics[width=\linewidth]{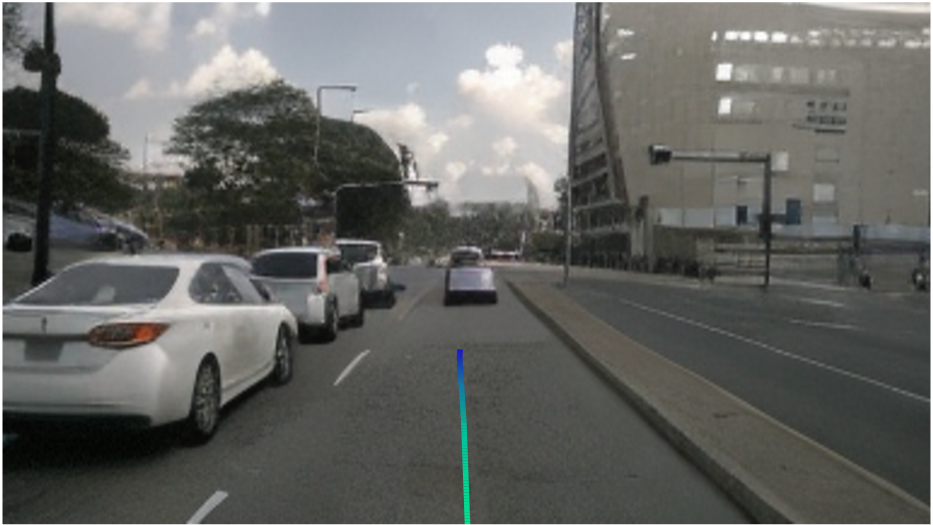}
    \caption{Planning with ego on centerline}
    \end{subfigure}
    \begin{subfigure}{0.48\columnwidth}
    \includegraphics[width=\linewidth]{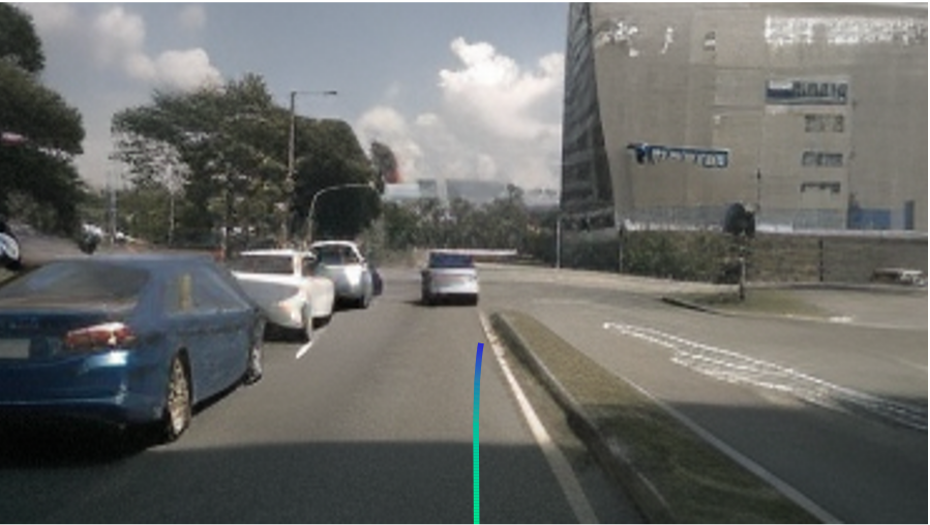}
    \caption{Planning with ego off centerline}
    \end{subfigure}
    \caption{
    \textbf{Ego vehicle's slight deviation from centerline causes motion planner to struggle generating reasonable trajectories.}
    We shift the ego location 0.5m to the right to create an out-of-domain case. (a) shows the reasonable trajectory prediction of the VAD~\cite{Jiang_2023_ICCV} method under normal data, and (b) shows the irrational trajectory when encountering out-of-distribution cases.}
    \label{fig:planner deviation}
    \vspace{-12pt}
\end{figure}

The emergence of end-to-end autonomous driving~\cite{tampuu2020survey, hu2023planning, Jiang_2023_ICCV} has recently garnered increasing attention.
These approaches take multi-sensor data as input and directly output planning results in a joint model, allowing for joint optimization of all modules. 
However, it is questionable whether an end-to-end planner trained purely on expert driving trajectories has sufficient generalization capabilities when faced with out-of-distribution (OOD) cases. 
As illustrated in Figure~\ref{fig:planner deviation}, when the ego vehicle's position deviates laterally from the center line, the end-to-end planner struggles to generate a reasonable trajectory.
To alleviate this problem, we propose improving the safety of autonomous driving by developing a predictive model that can foresee planner degradation before decision-making. 
This model, known as a world model~\cite{ha2018recurrent, Hafner2020Dream, lecun2022path}, is designed to predict future states based on current states and ego actions.
By visually envisioning the future in advance and obtaining feedback from different futures before actual decision-making, it can provide more rational planning, enhancing generalization and safety in end-to-end autonomous driving.

However, learning high-quality world models compatible with existing end-to-end autonomous driving models is challenging, despite successful attempts in game simulations~\cite{ha2018recurrent, Hafner2020Dream, schrittwieser2020mastering, hafner2021mastering} and laboratory robotics environments~\cite{ebert2018visual, Hafner2020Dream}.
Specifically, there are three main challenges:
(1) The driving world model requires modeling in high-resolution pixel space.
The previous low-resolution image~\cite{Hafner2020Dream} or vectorized state space~\cite{buesing2018learning} methods cannot effectively represent the numerous fine-grained or non-vectorizable events in the real world.
Moreover, vector space world models need extra vector annotations and suffer from state estimation noise of perception models.
(2) Generating multiview consistent videos is difficult.
Previous and concurrent works are limited to single view video~\cite{kim2021drivegan,wang2023drivedreamer,hu2023gaia} or multiview image generation~\cite{swerdlow2023street, yang2023bevcontrol, gao2023magicdrive}, leaving multiview video generation an open problem for comprehensive environment observation needed in autonomous driving.
(3) It is challenging to flexibly accommodate various heterogeneous conditions like changing weather, lighting, ego actions, and road/obstacle/vehicle layouts.

To address these challenges, we propose \methodname{}.
Inspired by latent video diffusion models~\cite{singer2022make, zhou2022magicvideo, blattmann2023align, esser2023structure}, we introduce multiview and temporal modeling for jointly generating multiple views and frames.
To further enhance multiview consistency, we propose factorizing the joint modeling to predict intermediate views conditioned on adjacent views, greatly improving consistency between views.
We also introduce a simple yet effective unified condition interface enabling flexible use of heterogeneous conditions like images, text, 3D layouts, and actions, greatly simplifying conditional generation.
Finally, building on the multiview world model, we explore end-to-end planning applications to enhance autonomous driving safety, as shown in Figure~\ref{fig:teaser}.
The main contributions of our work can be summarized as follows.

\begin{itemize}
    \item We propose \methodname{}, a multiview world model, which is capable of generating high-quality, controllable, and consistent \emph{multiview videos} in autonomous driving scenes.
    \item Extensive experiments on the nuScenes dataset showcase the leading video quality and controllability.
    \methodname{} also achieves superior multiview consistency, evaluated by a novel keypoint matching based metric.
    \item We are the first to explore the potential application of the world model in end-to-end planning for autonomous driving.
    We experimentally show that our method could enhance the overall soundness of planning and robustness in out-of-distribution situations.
\end{itemize}
\section{Related Works}
\label{sec:related works}

\subsection{Video Generation and Prediction}
Video generation aims to generate realistic video samples. Various generation methods have been proposed in the past, including VAE-based (Variational Autoencoder)~\cite{kingma2013auto, villegas2019high, franceschi2020stochastic, walker2021predicting}, GAN-based (generative adversarial networks)~\cite{tulyakov2018mocogan, fox2021stylevideogan, kim2021drivegan, yu2021generating, brooks2022generating, skorokhodov2022stylegan}, flow-based~\cite{kumar2019videoflow, dorkenwald2021stochastic} and auto-regressive models~\cite{weissenborn2019scaling, yan2021videogpt, ge2022long}. Notably, the recent success of diffusion-based models in the realm of image generation~\cite{nichol2022glide, rombach2022high, ruiz2023dreambooth} has ignited growing interest in applying diffusion models to the realm of video generation~\cite{harvey2022flexible,hoppe2022diffusion}. Diffusion-based methods have yielded significant enhancements in realism, controllability, and temporal consistency. Text-conditional video generation has garnered more attention due to its controllable generation, and a plethora of methods have emerged~\cite{ho2022imagen, singer2022make, zhou2022magicvideo, blattmann2023align, wu2023tune}. 

Video prediction can be regarded as a special form of generation, leveraging past observations to anticipate future frames~\cite{oh2015action, vondrick2016generating, denton2018stochastic, babaeizadeh2021fitvid, wu2021greedy, hoppe2022diffusion, voleti2022mcvd, hafner2023mastering}. Especially in autonomous driving, DriveGAN~\cite{kim2021drivegan} learns to simulate a driving scenario with vehicle control signals as its input. GAIA-1~\cite{hu2023gaia} and DriveDreamer~\cite{wang2023drivedreamer} further extend to action-conditional diffusion models, enhancing the controllability and realism of generated videos.
However, these previous works are limited to monocular videos and fail to comprehend the overall 3D surroundings. We have pioneered the generation of multiview videos, allowing for better integration with current BEV perception and planning models.

\subsection{World Model for Planning}
The world model~\cite{lecun2022path} learns a general representation of the world and predicts future world states resulting from a sequence of actions.
Learning world models in either game~\cite{ha2018recurrent, Hafner2020Dream, sekar2020planning, hafner2021mastering, reed2022generalist} or lab environments~\cite{Finn2016DeepVF, ebert2018visual, wu2023daydreamer} has been widely studied. Dreamer~\cite{Hafner2020Dream} learns a latent dynamics model from past experience to predict state values and actions in a latent space. It is capable of handling challenging visual control tasks in the DeepMind Control Suite~\cite{tassa2018deepmind}. DreamerV2~\cite{hafner2021mastering} improves upon Dreamer to achieve human-level performance on Atari games. DreamerV3~\cite{hafner2023mastering} uses larger networks and learns to obtain diamonds in Minecraft from scratch given sparse rewards, which is considered a long-standing challenge. DayDreamer~\cite{wu2023daydreamer} applies Dreamer~\cite{Hafner2020Dream} to training 4 robots online in the real world and solves locomotion and manipulation tasks without changing hyperparameters. Recently, learning world models in driving scenes has gained attention. MILE~\cite{hu2022model} employs a model-based imitation learning method to jointly learn a dynamics model and driving behaviour in CARLA~\cite{pmlr-v78-dosovitskiy17a}. The aforementioned works are limited to either simulators or well-controlled lab environments. In contrast, our world model can be integrated with existing end-to-end driving planners to improve planning performance in real-world scenes.

\begin{figure*}[th]
    \centering
    \includegraphics[width=\linewidth]{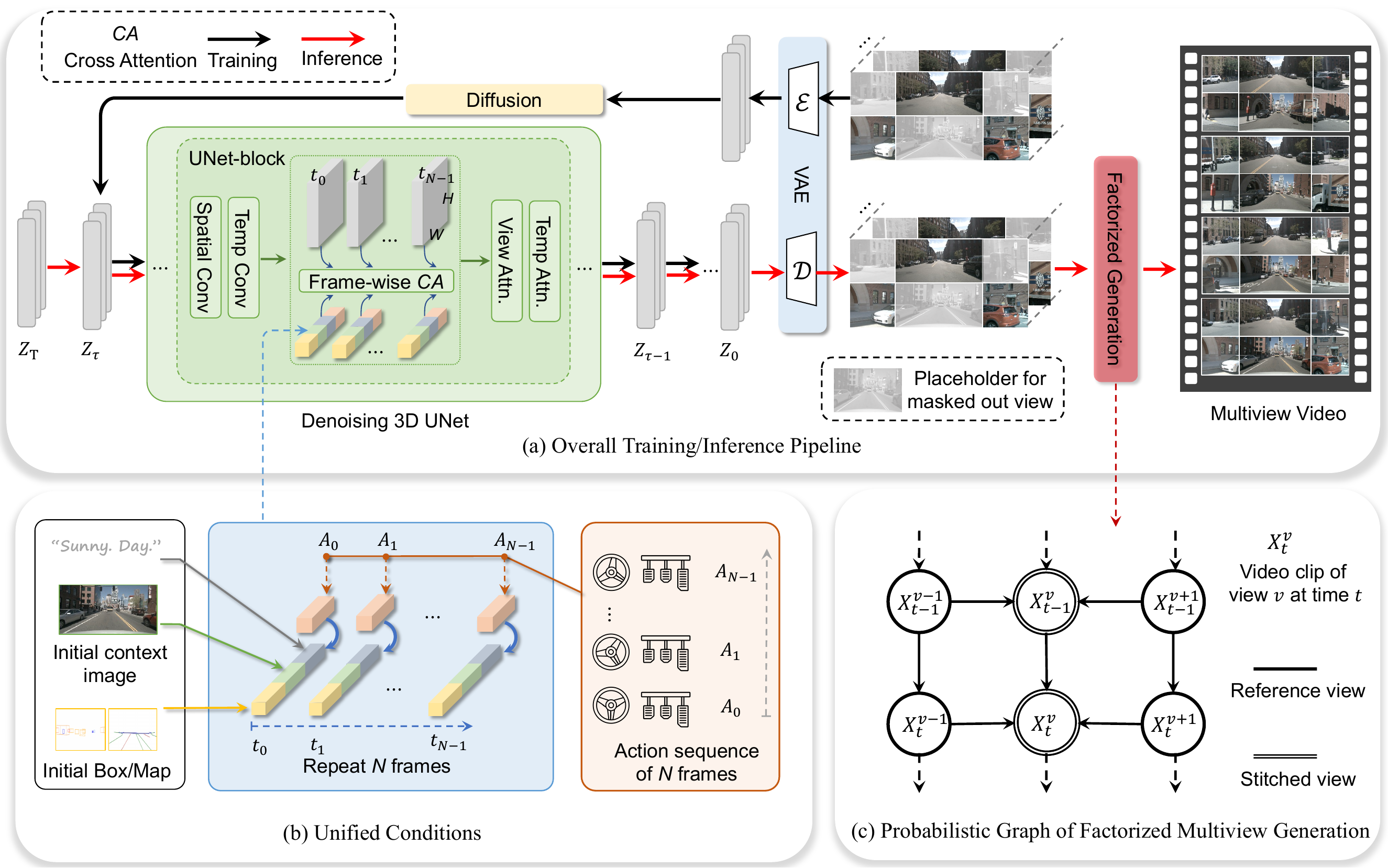}
    \caption{\textbf{Overview of the proposed framework.}
    \textbf{(a)} illustrates the training and inference pipeline of the proposed method.
    \textbf{(b)} visualizes the unified conditions leveraged to control the generation of multi-view video.
    \textbf{(c)} represents the probabilistic graph of factorized multiview generation. It takes the 3-view output from (a) as input to generate other views, enhancing the multi-view consistency.}
    \label{fig:framework}
    \vspace{-12pt}
\end{figure*}

\section{Multi-view Video Generation}
\label{sec:multiview}
In this section, we first present how to jointly model the multiple views and frames, which is presented in Sec~\ref{sec:joint_modeling}.
Then we enhance multiview consistency by factorizing the joint modeling in Sec~\ref{sec:factorization}.
Finally, Sec.~\ref{sec:unified_cond} elaborates on how we build a unified condition interface to integrate the multiple heterogeneous conditions.

\subsection{Joint Modeling of Multiview Video}
To jointly model multiview temporal data, we start with the well-studied image diffusion model and adapt it into multiview-temporal scenarios by introducing additional temporal layers and multiview layers.
In this subsection, we first present the overall formulation of joint modeling and elaborate on the temporal and multiview layers.
\label{sec:joint_modeling}
\paragraph{Formulation.}
We assume access to a dataset $p_\text{data}$ of multiview videos, such that $\x\in \mathbb{R}^{T\times K \times 3\times H\times W}$, $\x\sim p_\text{data}$ is a sequence of $T$ images with $K$ views, with height and width $H$ and $W$. 
Given encoded video latent representation $\mathcal{E}(\x)=\mathbf{z}\in\mathbb{R}^{T \cdot K \times C\times \hat{H} \times \hat{W}}$, diffused inputs $\mathbf{z_\tau} = \alpha_\tau\mathbf{z} + \sigma_\tau\beps$, $\beps\sim\cN(\0, \I)$, here $\alpha_\tau$ and $\sigma_\tau$ define a noise schedule parameterized by a diffusion time step $\tau$.
A denoising model $\f_{\theta,\phi,\psi}$ (parameterized by spatial parameters $\theta$, temporal parameters $\phi$ and multiview parameters $\psi$) receives the diffused $\mathbf{z}_\tau$ as input and is optimized by minimizing the denoising score matching objective
\begin{equation}
    \bE_{\mathbf{z}\sim p_\text{data},\tau\sim p_\tau,\beps\sim\cN(\0, \I)}[\|\y - \f_{\theta,\phi,\psi}(\mathbf{z}_\tau;\bc,\tau)\|_2^2],
    \label{eq:overall_formulation}
\end{equation}
where $\bc$ is the condition, 
and target $\y$ is the random noise $\beps$. $p_\tau$ is a uniform distribution over the diffusion time $\tau$.

\vspace{-3mm}
\paragraph{Temporal encoding layers.}
We first introduce \emph{temporal layers} to lift the pretrained image diffusion model into a temporal model.
The temporal encoding layer is attached after the 2D spatial layer in each block, following established practice in VideoLDM~\cite{blattmann2023align}.
The spatial layer encodes the latent $\mathbf{z}\in\mathbb{R}^{T \cdot K \times C\times \hat{H} \times \hat{W}}$ in a frame-wise and view-wise manner.
Afterward, we rearrange the latent to hold out the temporal dimension, denoted as $\texttt{(TK)CHW}\rightarrow\texttt{KCTHW}$, to apply the 3D convolution in spatio-temporal dimensions $\texttt{THW}$.
Then we arrange the latent to $\texttt{(KHW)TC}$ and apply standard multi-head self-attention to the temporal dimension, enhancing the temporal dependency.
The notation $\phi$ in Eq.~\ref{eq:overall_formulation} stands for the parameters of this part.

\vspace{-3mm}
\paragraph{Multiview encoding layers.}
To jointly model the multiple views, there must be information exchange between different views.
Thus we lift the single-view temporal model to a multi-view temporal model by introducing \emph{multiview encoding layers}.
In particular, we rearrange the latent as $\texttt{(KHW)TC} \rightarrow \texttt{(THW)KC}$ to hold out the view dimension.
Then a self-attention layer parameterized by $\psi$ in Eq.~\ref{eq:overall_formulation} is employed across the view dimension.
Such multiview attention allows all views to possess similar styles and consistent overall structure.

\vspace{-3mm}
\paragraph{Multiview temporal tuning.}
Given the powerful image diffusion models, we do not train the temporal multiview network from scratch.
Instead, we first train a standard image diffusion model with single-view image data and conditions, which corresponds to the parameter $\theta$ in Eq.~\ref{eq:overall_formulation}.
Then we freeze the parameters $\theta$ and fine-tune the additional temporal layers ($\phi$) and multiview layers ($\psi$) with video data.

\subsection{Factorization of Joint Multiview Modeling}
Although the joint distributions in Sec.~\ref{sec:joint_modeling} could yield similar styles between different views, it is hard to ensure strict consistency in their overlapped regions.
In this subsection, we introduce the distribution factorization to enhance multiview consistency.
We first present the formulation of factorization and then describe how it cooperates with the aforementioned joint modeling.
\label{sec:factorization}
\paragraph{Formulation.}
Let $\x_i$ denote the sample of $i$-th view, Sec.~\ref{sec:joint_modeling} essentially model the joint distribution $p(\x_{1, \dots, K})$, which can be into
\begin{equation}
p(\x_{1, \dots, K})  = p(\x_1)p(\x_2|\x_1) \dots p(\x_K | \x_1, \dots, \x_{K-1} ).
\label{eq:full_factorize}
\end{equation}
Eq.~\ref{eq:full_factorize} indicates that different views are generated in an autoregressive manner, where a new view is conditioned on existing views.
These conditional distributions can ensure better view consistency because new views are aware of the content in existing views.
However, such an autoregressive generation is inefficient, making such full factorization infeasible in practice.
\par
\begin{figure}
    \centering
    \includegraphics[width=\linewidth]{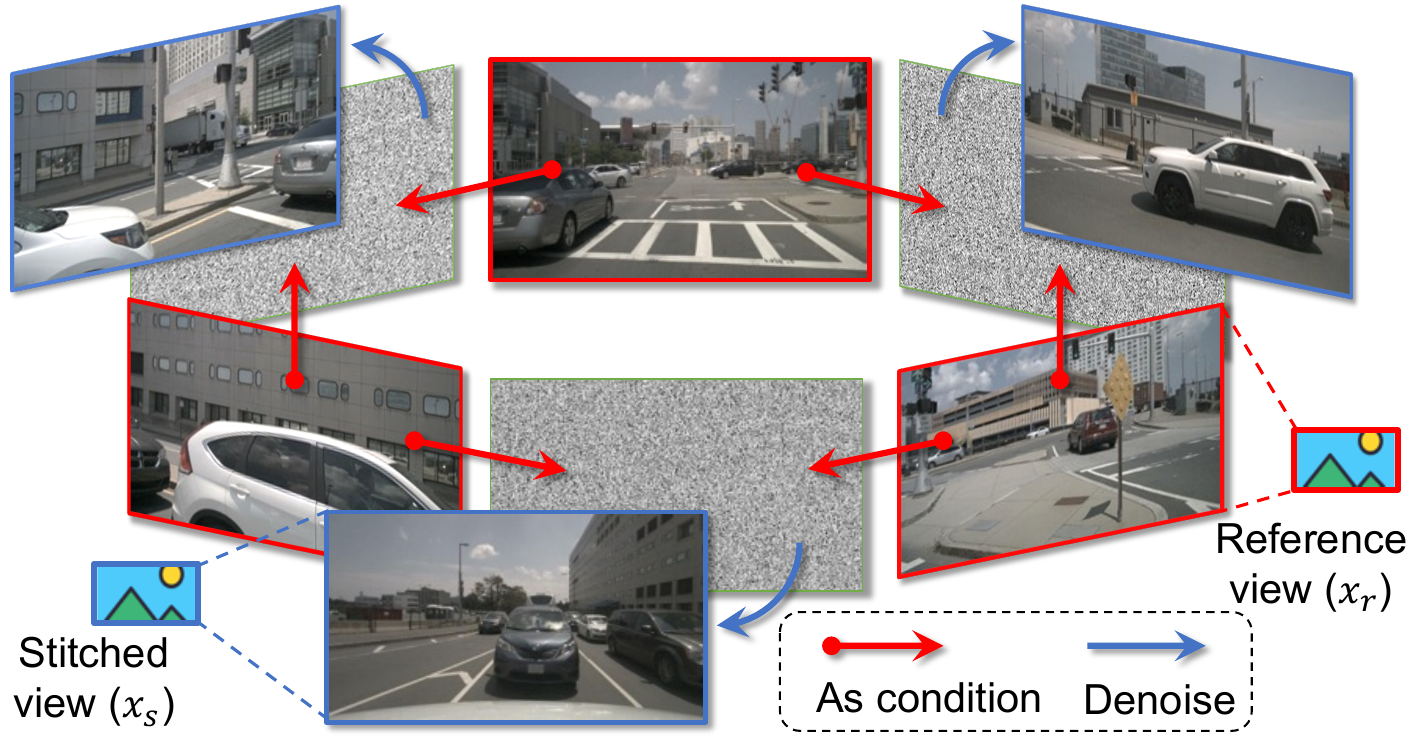}
    \caption{\textbf{Illustration of factorized multi-view generation.} We take the sensor layout in nuScenes as an example.}
    \label{fig:stitch}
    \vspace{-12pt}
\end{figure}

To simplify the modeling in Eq.~\ref{eq:full_factorize}, we partition all views into two types: \emph{reference views} $\x_r$ and \emph{stitched views} $\x_s$.
For example, in nuScenes, reference views can be the $\{\texttt{F, BL, BR} \}$\footnote{F: front, B: back, L: left, R: right.}, and stitched views can be $\{\texttt{FL, B, FR} \}$.
We use the term ``stitched'' because a stitched view appears to be ``stitched'' from its two neighboring reference views.
Views belonging to the same type do not overlap with each other, while different types of views may overlap.
This inspires us to first model the joint distribution of reference views.
Here the joint modeling is effective for those non-overlapped reference views since they do not necessitate strict consistency. 
Then the distribution of $\x_s$ is modeled as a conditional distribution conditioned on the $\x_r$.
Figure~\ref{fig:stitch} illustrates the basic concept of multiview factorization in nuScenes.
In this sense, we simplify Eq.~\ref{eq:full_factorize} into 
\begin{equation}
    p(\x) = p(\x_s, \x_r) = p(\x_r)p(\x_s | \x_r).
    \label{eq:factorization1}
\end{equation}
Considering the temporal coherence, we incorporate previous frames as additional conditions. The Eq.~\ref{eq:factorization1} can be re-written as 
\begin{equation}
    p(\x) = p(\x_s, \x_r | \x_{pre}) = p(\x_r | \x_{pre} )p(\x_s | \x_r, \x_{pre}),
    \label{eq:factorization2}
\end{equation}
where $\x_{pre}$ is context frames (e.g., the last two frames) from previously generated video clips.
The distribution of reference views $p(\x_r|\x_{pre})$ is implemented by the pipeline in Sec.~\ref{sec:joint_modeling}. As for $p(\x_s | \x_r, \x_{pre})$, we adopt the similar pipeline but incorporate neighboring reference views as an additional condition as Figure~\ref{fig:stitch} shows.
We introduce how to use conditions in the following subsection.

\subsection{Unified Conditional Generation}
\label{sec:unified_cond}
Due to the great complexity of the real world, the world model needs to leverage multiple heterogeneous conditions.
In our case, we utilize initial context frames, text descriptions, ego actions, 3D boxes, BEV maps, and reference views.
More conditions can be further included for better controllability. 
Developing specialized interface for each one is time-consuming and inflexible to incorporate more conditions.
To address this issue, we introduce a unified condition interface, which is simple yet effective in integrating multiple heterogeneous conditions.
In the following, we first introduce how we encode each condition, and then describe the unified condition interface.

\vspace{-3mm}
\paragraph{Image condition.}
We treat initial context frames (i.e., the first frame of a clip) and reference views as image conditions.
A given image condition $\mathbf{I}\in \mathbb{R}^{3\times H \times W}$ is encoded and flattened to a sequence of $d$-dimension embeddings $\mathbf{i} = (i_1,i_2,...,i_n) \in \mathbb{R}^{n\times d}$, using ConvNeXt as encoder~\cite{liu2022convnet}.
Embeddings from different images are concatenated in the first dimension of $n$.
\vspace{-5mm}
\paragraph{Layout condition.}
Layout condition refers to 3D boxes, HD maps, and BEV segmentation.
For simplicity, we project the 3D boxes and HD maps into a 2D perspective view.
In this way, we leverage the same strategy with image condition encoding to encode the layout condition, resulting in a sequence of embeddings $\mathbf{l} = (l_1,l_2,...,l_k) \in \mathbb{R}^{k\times d}$.
$k$ is the total number of embeddings from the projected layouts and BEV segmentation.
\vspace{-5mm}
\paragraph{Text condition.}
We follow the convention of diffusion models to adopt a pre-trained CLIP~\cite{radford2021learning} as the text encoder. Specifically, we combine view information, weather, and light to derive a text description. The embeddings are denoted as $\mathbf{e} = (e_1,e_2,...,e_m) \in \mathbb{R}^{m\times d}$.
\vspace{-5mm}
\paragraph{Action condition.}
Action conditions are indispensable for the world model to generate the future.
To be compatible with the existing planning methods~\cite{Jiang_2023_ICCV}, we define the action in a time step as $(\Delta x, \Delta y)$, which represents the movement of ego location to the next time step. 
We use an MLP to map the action into a $d$-dimension embedding $\mathbf{a} \in \mathbb{R}^{2 \times d}$.
\vspace{-5mm}
\paragraph{A unified condition interface.}
So far, all the conditions are mapped into $d$-dimension feature space.
We take the concatenation of required embeddings as input for the denoising UNet.
Taking action-based joint video generation as an example, this allows us to utilize the initial context images, initial layout, text description, and frame-wise action sequence.
So we have unified condition embeddings in a certain time $t$ as
\vspace{-2mm}
\begin{equation}
   \mathbf{c}_t = [\mathbf{i}_0, \mathbf{l}_0, \mathbf{e}_0, \mathbf{a}_t] \in \mathbb{R}^{(n+k+m+2)\times d},  
   \label{eq:action_cond}
   \vspace{-2mm}
\end{equation}
where subscript $t$ stands for the $t$-th \emph{generated} frame and subscript $0$ stands for the current \emph{real} frame. 
We emphasize that such a combination of different conditions offers a unified interface and can be adjusted by the request.
Finally, $\mathbf{c_t}$ interacts with the latent $\mathbf{z_t}$ in 3D UNet by cross attention in a frame-wise manner (Figure~\ref{fig:framework} (a)).
    
\section{World Model for End-to-End Planning}
\label{sec:rollout}
\begin{figure}[t]
    \centering
    \includegraphics[width=\linewidth]{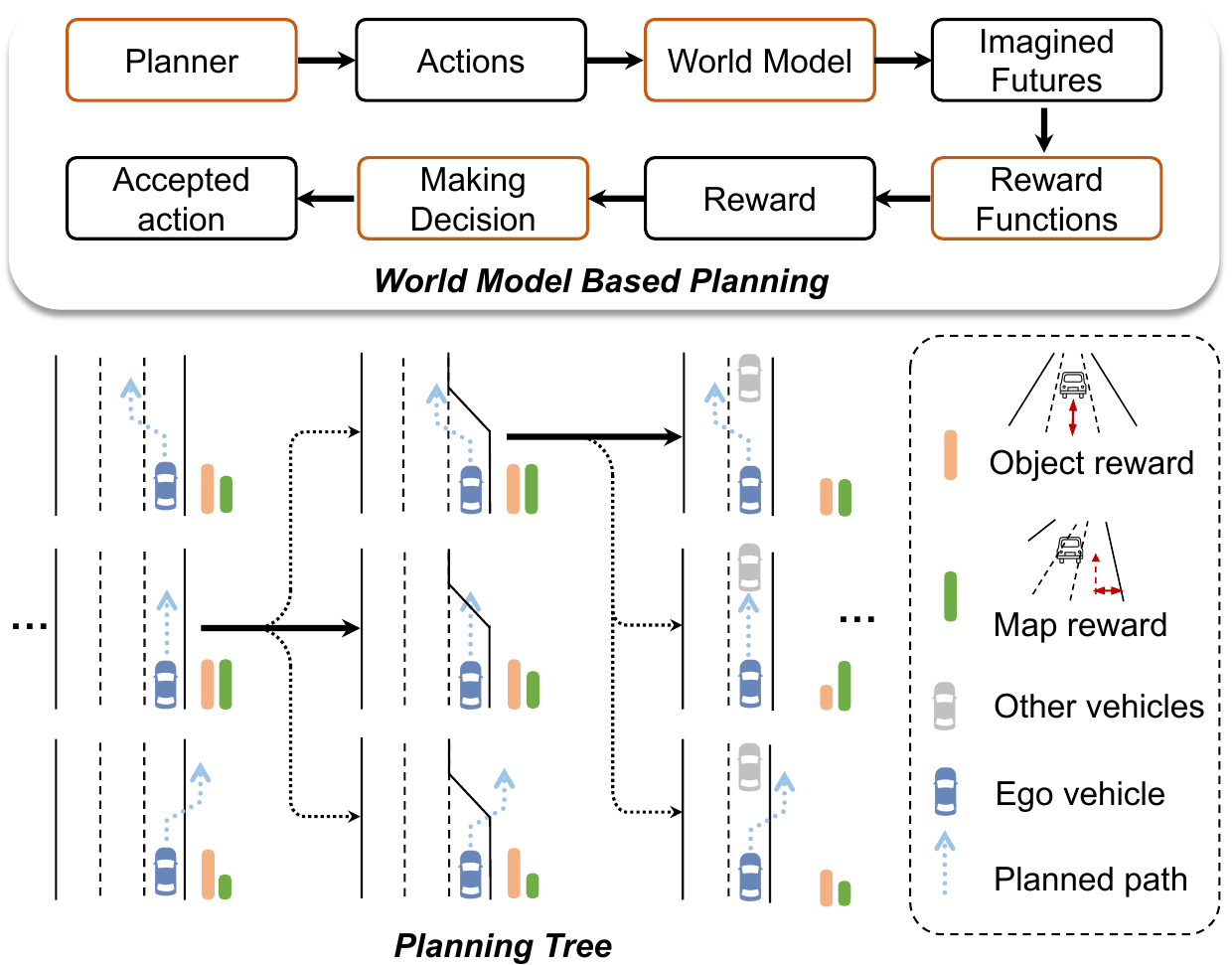}
    \caption{\textbf{End-to-end planning pipeline with our world model.} We display the components of our planning pipeline at the top and illustrate the decision-making process in the planning tree using image-based rewards at the bottom.}
    \label{fig:planner_pipeline}
    \vspace{-12pt}
\end{figure}

Blindly planning actions without anticipating consequences is dangerous. 
Leveraging our world model enables comprehensive evaluation of possible futures for safer planning. 
In this section, we explore end-to-end planning using the world model for autonomous driving, an uncharted area.

\subsection{Tree-based Rollout with Actions}

We describe planning with world models in this section. At each time step, we leverage the world model to generate predicted future scenarios for trajectory candidates sampled from the planner, evaluate the futures using an image-based reward function, and select the optimal trajectory to extend the planning tree.

As shown in Figure~\ref{fig:planner_pipeline}, we define the planning tree as a series of predicted ego trajectories that evolve over time.
For each time, the real multiview images can be captured by the camera.
The pre-trained planner takes the real multiview images as input and samples possible trajectory candidates.
To be compatible with the input of mainstream planner, we define its action $\mathbf{a_t}$ at time $t$ as $(x_{t+1} - x_{t}, y_{t+1} - y_{t})$ for each trajectory, where $x_t$ and $y_t$ are the ego locations at time $t$.
Given the actions, we adopt the condition combination in Eq.~\ref{eq:action_cond} for the video generation.
After the generation, we leverage an image-based reward function to choose the optimal trajectory as the decision.
Such a generation-decision process can be repeated to form a tree-based rollout.
\subsection{Image-based Reward Function}
After generating the future videos for planned trajectories, reward functions are required to evaluate the soundness of the multiple futures.
\par
We first get the rewards from perception results.
Particularly, we utilize image-based 3D object detector~\cite{li2022bevformer} and online HDMap predictor~\cite{liao2022maptr} to obtain the perception results on the generated videos.
Then we define map reward and object reward, inspired by traditional planner~\cite{casas2021mp3,Jiang_2023_ICCV}.
The map reward includes two factors, distance away from the curb, encouraging the ego vehicle to stay in the correct drivable area, and centerline consistency, preventing ego from frequently changing lanes and deviating from the lane in the lateral direction.
The object reward means the distance away from other road users in longitudinal and lateral directions. This reward avoids the collision between the ego vehicle and other road users. The total reward is defined as the product of the object reward and the map reward.
We finally select the ego prediction with the maximum reward. Then the planning tree forwards to the next timestamp and plans the subsequent trajectory iteratively.
\par
Since the proposed world model operates in pixel space, it can further get rewards from the \emph{non-vectorized} representation to handle more general cases.
For example, the sprayed water from the sprinkler and damaged road surface are hard to be vectorized by the supervised perception models, while the world model trained from massive unlabeled data could generate such cases in pixel space.
Leveraging the recent powerful foundational models such as GPT-4V, the planning process can get more comprehensive rewards from the non-vectorized representation. In the appendix, we showcase some typical examples.

\section{Experiments}
\label{sec:experiment}
\subsection{Setup}
\label{sec:setup}
\noindent\textbf{Dataset.} 
We adopt the nuScenes~\cite{caesar2020nuscenes} dataset for experiments, which is one of the most popular datasets for  3D perception and planning.
It comprises a total of 700 training videos and 150 validation videos. Each video includes around 20 seconds captured by six surround-view cameras.

\noindent\textbf{Training scheme.}
We crop and resize the original image from 1600 $\times$ 900 to 384 $\times$ 192. 
Our model is initialized with Stable Diffusion checkpoints~\cite{rombach2022high}. 
All experiments are conducted on A40 (48GB) GPUs. 
For additional details, please refer to the appendix.

\noindent\textbf{Quality evaluation.}
To evaluate the quality of the generated video, we utilize FID (Frechet Inception Distance)~\cite{heusel2017gans} and FVD (Frechet Video Distance)~\cite{unterthiner2018towards} as the main metrics.

\setlength{\tabcolsep}{3pt}
\setlength{\doublerulesep}{2\arrayrulewidth}
\renewcommand{\arraystretch}{1.1}
\renewcommand{\multirowsetup}{\centering}

\begin{table*}[t]
\subfloat[
\textbf{Generation quality}. 
\label{tab:quality}
]{
\begin{minipage}{0.45\linewidth}
{
\small
\begin{tabular}{l|cc|cc}
    \toprule
    Method & Multi-view & Video & FID$\downarrow$ & FVD$\downarrow$ \\
    \midrule
    BEVGen~\cite{swerdlow2023street} & \checkmark & & 25.54 &-  \\
    BEVControl~\cite{yang2023bevcontrol} & \checkmark & &24.85 & - \\
    MagicDrive~\cite{gao2023magicdrive} &  \checkmark & &16.20 & - \\
    Ours & \checkmark &  & \textbf{12.99}  & - \\
    \midrule
    DriveGAN~\cite{kim2021drivegan} &  & \checkmark & 73.4 & 502.3 \\
    DriveDreamer~\cite{wang2023drivedreamer} &  & \checkmark & 52.6 & 452.0\\
    Ours & \checkmark & \checkmark & \textbf{15.8}  & \textbf{122.7}  \\
    \bottomrule
  \end{tabular}
}
\end{minipage}
}
\hspace{1em}
\subfloat[
\textbf{Generation controllability}. 
\label{tab:controllability}
]{
\begin{minipage}{0.5\linewidth}
{
\small 
\begin{tabular}{l|cccc}
    \toprule
     Method & mAP$_\text{obj}$ $\uparrow$ & mAP$_\text{map}$ $\uparrow$ & mIoU$_\text{fg}$ $\uparrow$ & mIoU$_\text{bg}$ $\uparrow$ \\
    \midrule
    \textcolor{gray} {GT} & \textcolor{gray}{37.78} &\textcolor{gray}{59.30} &  \textcolor{gray}{36.08}&\textcolor{gray}{72.36}\\
    \midrule
    BEVGen~\cite{swerdlow2023street} & - &- &5.89 & 50.20  \\
    LayoutDiffusion~\cite{zheng2023layoutdiffusion} & 3.68 & - & 15.51 & 35.31 \\
    GLIGEN~\cite{li2023gligen} & 15.42 & -&22.02 & 38.12 \\ 
    BEVControl~\cite{yang2023bevcontrol} & 19.64 & -&26.80 & 60.80 \\
    MagicDrive~\cite{gao2023magicdrive} & 12.30 & - & 27.01 & 61.05 \\
    Ours & \textbf{20.66}& \textbf{37.68}& \textbf{27.19} & \textbf{65.07} \\
    \bottomrule
  \end{tabular}
}
\end{minipage}
}
\vspace{-8pt}
\caption{\textbf{Multi-view video generation performance on nuScenes.} For each task, we test the corresponding models trained on the nuScenes training set. Our \methodname{} surpasses all other methods in both quality and controllability evaluation.}
\label{table:video_quality}
\end{table*}
\begin{table*}[t]
\subfloat[
\textbf{Ablations of unified condition.}
\label{tab:unified_condition}
]{
\begin{minipage}{0.3\linewidth}
{
\small 
\resizebox{\columnwidth}{!}{
\begin{tabular}{cc|ccc}
\toprule
Temp emb. & Layout Cond. & FID$\downarrow$ & FVD$\downarrow$ & KPM(\%)$\uparrow$   \\
\midrule
 \checkmark &  &20.3 & 212.5 &31.5 \\
  &  \checkmark & 18.9 & 153.8 & 44.6   \\
 \checkmark & \checkmark &\textbf{15.8} & \textbf{122.7}& \textbf{45.8}   \\
\bottomrule
\end{tabular}
}}
\end{minipage}
}
\hspace{1em}
\subfloat[
\textbf{Ablations of multiview temporal tuning.}
\label{tab:attn}
]{
\begin{minipage}{0.3\linewidth}
{
\small 
\resizebox{\columnwidth}{!}{
\begin{tabular}{cc|ccc}
\toprule
Temp Layers & View Layers & FID$\downarrow$ & FVD$\downarrow$ & KPM(\%)$\uparrow$   \\
\midrule
& & 23.3 &228.5 & 40.8 \\
 \checkmark &  & 16.2 & 127.1& 40.9 \\
 \checkmark & \checkmark &\textbf{15.8} & \textbf{122.7}& \textbf{45.8}  \\
\bottomrule
\end{tabular}
}}
\end{minipage}
}
\hspace{1em}
\subfloat[
\textbf{Ablations of factorized generation.} 
\label{tab:stitching}
]{
\begin{minipage}{0.3\linewidth}
{
\small
\resizebox{\columnwidth}{!}{
\begin{tabular}{l|ccc}
\toprule
{Method} & KPM(\%)$\uparrow$ & FVD$\downarrow$ & FID$\downarrow$\\
\midrule
Joint Modeling & 45.8 &   122.7 & \textbf{15.8}  \\
Factorized Generation &\textbf{94.4} & \textbf{116.6}  & 16.4 \\

\bottomrule
\end{tabular}}
}
\end{minipage}
}
\vspace{-8pt}
\caption{\textbf{Ablations of the components in model design.} The experiments are conducted under the layout-based video generation (See model variants in Sec.~\ref{sec:setup}) from nuScenes validation set.}
\label{table:video_ablation}
\vspace{-12pt}
\end{table*}

\noindent\textbf{Multiview consistency evaluation.}
\label{sec:kpm}
We introduced a novel metric, the Key Points Matching (KPM) score, to evaluate multi-view consistency.
This metric utilizes a pre-trained matching model~\cite{sun2021loftr} to calculate the average number of matching key points, thereby quantifying the KPM score.
In the calculation process, for each image, we first compute the number of matched key points between the current view and its two adjacent views.
Subsequently, we calculated the ratio between the number of matched points in generated data and the number of matched points in real data.
Finally, we averaged these ratios across all generated images to obtain the KPM score.
In practice, we uniformly selected 8 frames per scene in the validation set to calculate KPM.

\noindent\textbf{Controllability evaluation.}
To assess the controllability of video content generation, we evaluate generated images using pre-trained perception models. Following the previous method~\cite{yang2023bevcontrol}, we adopted CVT~\cite{zhou2022cross} for foreground and background segmentation. Additionally, we evaluate 3D object detection~\cite{wang2023exploring} and online map construction~\cite{liao2022maptr}.

\noindent\textbf{Planning evaluation.}
We follow open-loop evaluation metrics~\cite{hu2023planning,Jiang_2023_ICCV} for end-to-end planning, including L2 distance from the GT trajectory and the object collision rate.

\noindent\textbf{Model variants.}
We support action-based video generation and layout-based video generation. The former gives the ego action of each frame as the condition, while the latter gives the layout (3D box, map information) of each frame.

\subsection{Main Results of Multi-view Video Generation}
We first demonstrate our superior generation quality and controllability. 
Here the generation is conditioned on frame-wise 3D layouts.
Our model is trained in nuScenes \textbf{\emph{train}} split, and evaluated with the conditions in \textbf{\emph{val}} split.

\vspace{-3mm}
\paragraph{Generation quality.}
Since we are the first one to explore multi-view video generation, we make separate comparisons with previous methods in \emph{multi-view images} and \emph{single-view videos}, respectively.
For multi-view image generation, we remove the temporal layers in Sec.~\ref{sec:joint_modeling}.
Table~\ref{tab:quality} showcases the main results.
In single-view image generation, we achieve \textbf{12.99} FID, achieving a significant improvement over previous methods.
For video generation, our method exhibits a significant quality improvement compared to past single-view video generation methods, achieving \textbf{15.8} FID and  \textbf{122.7} FVD.
Additionally, our method is the first work to generate consistent multi-view videos, which is quantitatively demonstrated in Sec.~\ref{sec:ablation}.

\vspace{-3mm}
\paragraph{Controllability.}
In Table~\ref{tab:controllability}, we examine the controllability of our method on the nuScenes validation set.
For foreground controllability, we evaluate the performance of 3D object detection on the generated multiview videos, reporting mAP$_\text{obj}$.
Additionally, we segment the foreground on the BEV layouts, reporting mIoU$_\text{fg}$.
Regarding background control, we report the mIoU of road segmentation.
Furthermore, we evaluate mAP$_\text{map}$ for HDMap performance.
This superior controllability highlights the effectiveness of the unified condition interface (Sec.~\ref{sec:unified_cond}) and demonstrates the potential of the world model as a neural simulator.

\subsection{Ablation Study for Multiview Video Generation}
\label{sec:ablation}
To validate the effectiveness of our design decisions, we conduct ablation studies on the key features of the model, as illustrated in Table~\ref{table:video_ablation}.
The experiments are conducted under layout-based video generation.

\begin{figure*}
    \centering
    \includegraphics[width=1\textwidth]{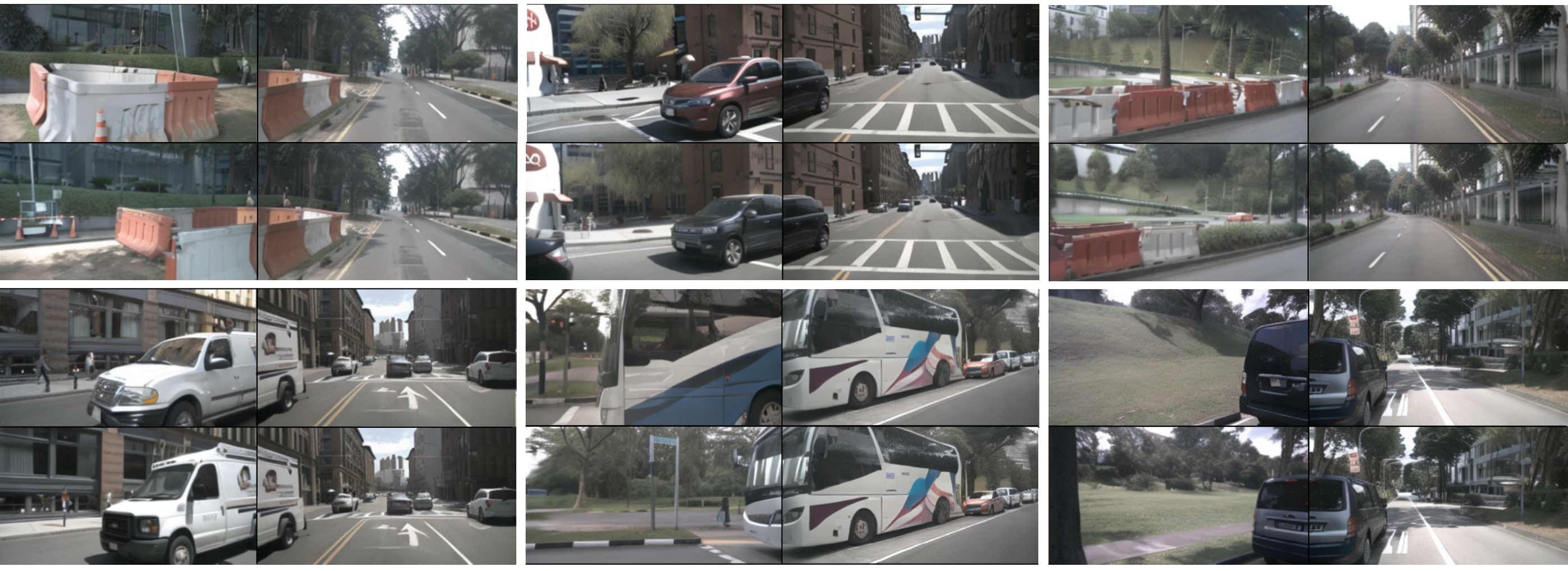}
    \captionof{figure}{\textbf{Qualitative results of factorized multiview generation}. For each compared pair, the upper row is generated without factorization, and the lower row is generated with factorization.}
    \label{fig:stitch_vis}
    \vspace{-10pt}
\end{figure*}

\begin{figure*}[ht]
    \centering
    \includegraphics[width=\linewidth]{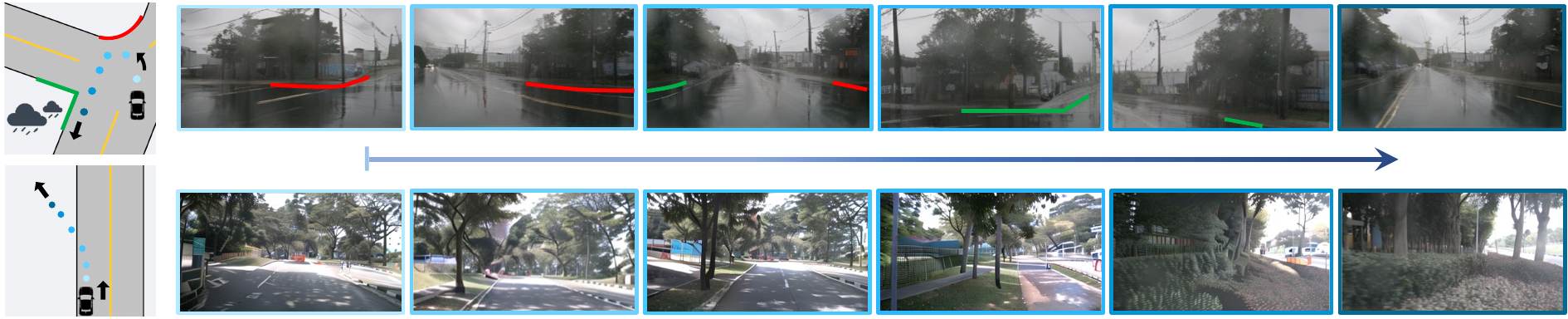}
    \caption{\textbf{Counterfactual events generation.} Top: turning around at a T-shape intersection on a rainy day. Note that our training set does not contain any turning-around samples. Bottom: running over a non-drivable area.}
    \label{fig:counter_factual}
    \vspace{-10pt}
\end{figure*}

\vspace{-3mm}
\paragraph{Unified condition.}
In Table~\ref{tab:unified_condition}, we find that the layout condition has a significant impact on the model's ability, improving both the quality and consistency of the generated videos. Additionally, temporal embedding can enhance the quality of the generated videos.  

\vspace{-3mm}
\paragraph{Model design.}
In Table~\ref{tab:attn}, we explore the role of the temporal and view layers in multiview temporal tuning. The experiment shows that the temporal and view layers Simply adopting the multiview layer without factorization (Sec.~\ref{sec:factorization}) slightly improve the KPM.

\vspace{-3mm}
\paragraph{Factorized multiview generation.}
As indicated in Table~\ref{tab:stitching}, factorized generation notably improves the consistency among multiple views, increasing from 45.8\% to 94.4\%, in contrast to joint modeling. This enhancement is achieved while ensuring the quality of both images and videos. Qualitative results are illustrated in Figure~\ref{fig:stitch_vis}.

\subsection{Exploring Planning with World Model}
In this subsection, we explore the application of the world model in end-to-end planning, which is under-explored in recent works for autonomous driving.
Our attempts lie in two aspects.
(1) We first demonstrate that evaluating the generated futures is helpful in planning.
(2) Then we showcase that the world model can be leveraged to improve planning in some out-of-domain cases.

 \begin{table}[hb]
\begin{center}
\resizebox{\columnwidth}{!}{
\begin{tabular}{l|cccc|cccc}
\toprule
\multirow{2}{*}{Method} &
\multicolumn{4}{c|}{L2 (m) $\downarrow$} & 
\multicolumn{4}{c}{Collision (\%) $\downarrow$} \\
& 1s & 2s & 3s & \cellcolor{gray!30}Avg. & 1s & 2s & 3s & \cellcolor{gray!30}Avg. \\
\midrule
\textcolor{gray}{VAD (GT cmd)}  & \textcolor{gray}{0.41} & \textcolor{gray}{0.70} & \textcolor{gray}{1.05} & \cellcolor{gray!30}\textcolor{gray}{0.72} & \textcolor{gray}{0.07} & \textcolor{gray}{0.17} & \textcolor{gray}{0.41} & \cellcolor{gray!30}\textcolor{gray}{0.22} \\
\midrule
VAD (rand cmd)&0.51 &0.97 & 1.57 & \cellcolor{gray!30}1.02 &0.34 &0.74 &1.72 & \cellcolor{gray!30}0.93 \\
Ours &\textbf{0.43}&\textbf{0.77} &\textbf{1.20} & \cellcolor{gray!30}\textbf{0.80}&\textbf{0.10} &\textbf{0.21} &\textbf{0.48} &\cellcolor{gray!30}\textbf{0.26}\\
\bottomrule
\end{tabular}}
\end{center}
\vspace{-8pt}
\caption{\textbf{Planning performance on nuScenes.} Instead of using the ground truth driving command, we use our tree-based planning to select the best out of three commands.
}
\label{tab:sota-plan}
\vspace{-12pt}
\end{table}

\begin{table}[ht]
\begin{center}
\resizebox{\columnwidth}{!}{
\begin{tabular}{cc|cccc|cccc}
\toprule
Map &Object  &
\multicolumn{4}{c|}{L2 (m) $\downarrow$} & 
\multicolumn{4}{c}{Collision (\%) $\downarrow$} \\
Reward&Reward& 1s & 2s & 3s & \cellcolor{gray!30}Avg. & 1s & 2s & 3s & \cellcolor{gray!30}Avg. \\
\midrule
& &0.51 &0.97 & 1.57 & \cellcolor{gray!30}1.02 &0.34 &0.74 &1.72 & \cellcolor{gray!30}0.93 \\
\checkmark& &0.45&0.82 &1.29 & \cellcolor{gray!30}0.85 &0.12 &0.33 &0.72 &\cellcolor{gray!30}0.39\\
&\checkmark &0.43&0.77 &1.20 & \cellcolor{gray!30}0.80 &0.12 &0.21 &0.48& \cellcolor{gray!30}0.27\\
\checkmark&\checkmark &\textbf{0.43}&\textbf{0.77} &\textbf{1.20} & \cellcolor{gray!30}\textbf{0.80}&\textbf{0.10} &\textbf{0.21} &\textbf{0.48} &\cellcolor{gray!30}\textbf{0.26}\\
\bottomrule
\end{tabular}}
\end{center}
\vspace{-10pt}
\caption{\textbf{Image-based reward function design.} We use two sub-rewards, map reward and object reward.
}
\label{tab:abl-plan}
\vspace{-12pt}
\end{table}
\vspace{-3mm}
\paragraph{Tree-based planning.}
We conduct the experiments to show the performance of our tree-based planning.
Instead of using the ground truth driving command, we sample planned trajectories from VAD according to the three commands ``Go straight", ``Turn left", and ``Turn right". Then the sampled actions are used for our tree-based planning (Sec.~\ref{sec:rollout}).
As shown in Table~\ref{tab:sota-plan}, our tree-based planner outperforms random driving commands and even achieves performance close to the GT command. Besides, in Table~\ref{tab:abl-plan}, we ablate two adopted rewards and the results indicate that the combined reward outperforms each sub-reward, particularly in terms of the object collision metric. 

\begin{table}[th]
\begin{center}
\resizebox{\columnwidth}{!}{
\begin{tabular}{cc|cccc|cccc}
\toprule
\multirow{2}{*}{OOD} &\multirow{2}{*}{World Model f.t.} &
\multicolumn{4}{c|}{L2 (m) $\downarrow$} & 
\multicolumn{4}{c}{Collision (\%) $\downarrow$} \\
&& 1s & 2s & 3s & \cellcolor{gray!30}Avg. & 1s & 2s & 3s & \cellcolor{gray!30}Avg. \\
\midrule
& & 0.41 & 0.70 & 1.05 & \cellcolor{gray!30}0.72 & 0.07 & 0.17 & 0.41 & \cellcolor{gray!30}0.22 \\
 \checkmark & &0.73 &0.99 &1.33 &\cellcolor{gray!30}1.02 &1.25 &1.62 &1.91 & \cellcolor{gray!30}1.59\\
 \checkmark & \checkmark &0.50 &0.79 &1.17 &\cellcolor{gray!30}0.82 &0.72 &0.84 &1.16 & \cellcolor{gray!30}0.91\\

\bottomrule
\end{tabular}}
\end{center}
\vspace{-8pt}
\caption{\textbf{Out-of-domain planning.} We define OOD location with a lateral deviation of 0.5 meters from the ego vehicle.
}
\label{tab:ood-plan}
\vspace{-12pt}
\end{table}
\vspace{-3mm}
\paragraph{Recovery from OOD ego deviation.}
Using our world model, we can simulate the out-of-domain ego locations in pixel space. In particular, we shift the ego location laterally by 0.5 meters, like the right one in Figure~\ref{fig:planner deviation}. 
In this situation, the performance of the existing end-to-end planner VAD~\cite{Jiang_2023_ICCV} undergoes a significant decrease, shown in Table~\ref{tab:ood-plan}. 
To alleviate the problem, we fine-tune the planner with generated video supervised by the trajectory that the ego-vehicle drives back to the lane. 
Learning from these OOD data, the performance of the planner can be better and near normal levels.\par

\subsection{Counterfactual Events}
Given an initial observation and the action, our \methodname{} can generate 
counterfactual events, such as turning around and running over non-drivable areas (Figure~\ref{fig:counter_factual}), which are significantly different from the training data.
The ability to generate such counterfactual data reveals again that our \methodname{} has the potential to foresee and handle out-of-domain cases.
\section{Conclusion}
\label{sec:conlusion}
We introduce \methodname{}, the first multiview world model for autonomous driving. 
Our method exhibits the capability to generate high-quality and consistent multiview videos under diverse conditions, leveraging information from textual descriptors, layouts, or ego actions to control video generation. The introduced factorized generation significantly enhances spatial consistency across various views.
Besides, extensive experiments on nuScenes dataset show that our method could enhance the overall soundness of planning and robustness in out-of-distribution situations.

{
    \small
    \bibliographystyle{ieeenat_fullname}
    \bibliography{main}
}

\clearpage
\clearpage
\setcounter{page}{12}
\maketitlesupplementary

\appendix

\section{Qualitative Results}
In this section, we present qualitative examples that demonstrate the performance of our model.
For the full set of video results generated by our model, please see our anonymized project page at \url{https://drive-wm.github.io}.

\subsection{Generation of Multiple Plausible Futures}
\begin{figure*}[ht]
    \includegraphics[width=\linewidth]{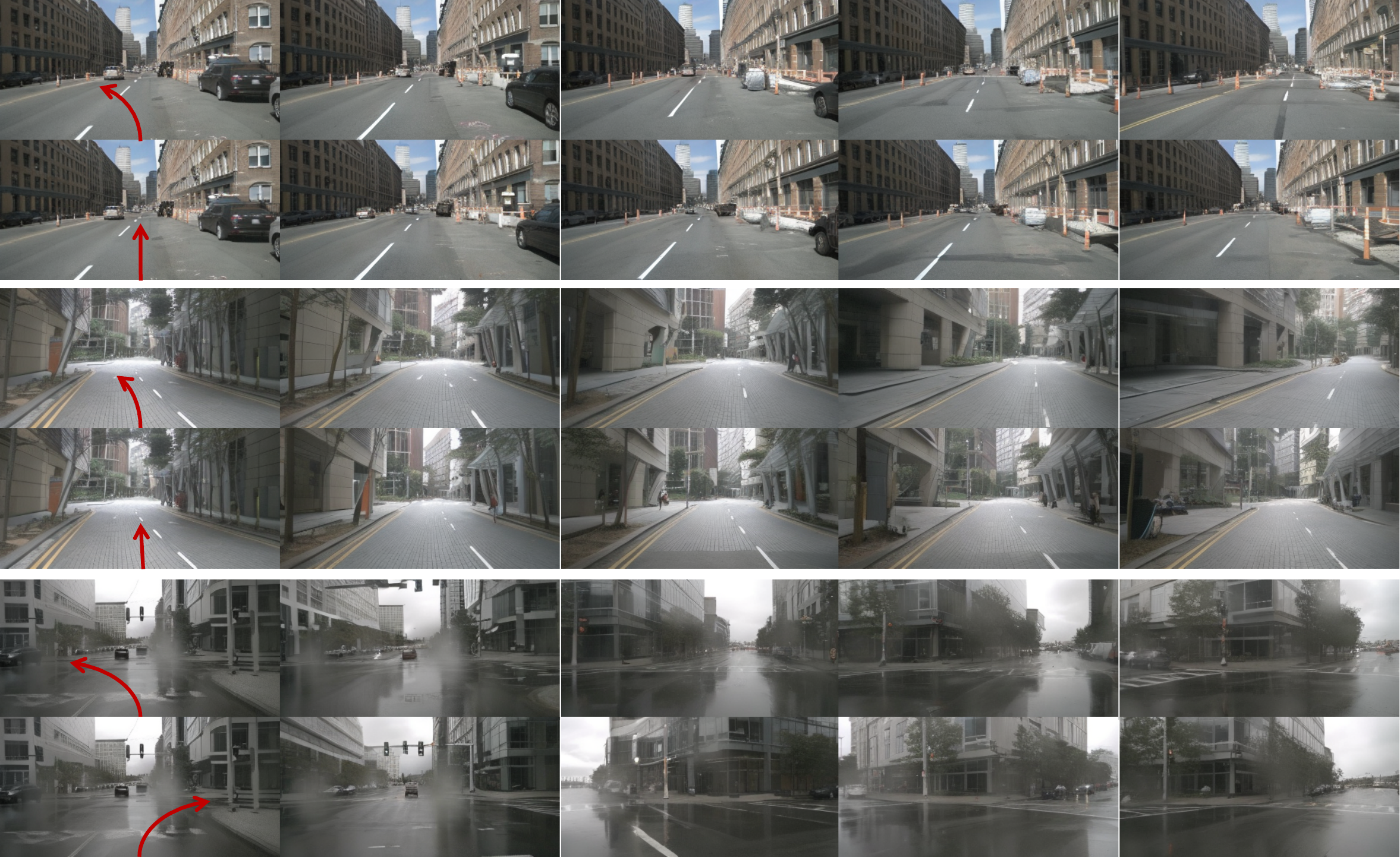}
    \caption{\textbf{Generation of multiple plausible futures based on the planning.} Here we only show the front-view videos for better illustration. The video samples are generated based on the first frames from the nuScenes \emph{val} set. In the first row, we show examples of lane changing to the left and straight ahead. In the second row, we show cases of driving towards the roadside and driving straight ahead. In the last row, we present examples of making left and right turns at the intersection.}
    \label{fig:multi_future}
\end{figure*}

\methodname{} can predict diverse future outcomes according to maneuvers from planners, as shown in Figure~\ref{fig:multi_future}. Based on plans from VAD~\cite{Jiang_2023_ICCV}, \methodname{} forecasts multiple plausible futures consistent with the initial observation.
We generate video samples on the nuScenes~\cite{caesar2020nuscenes} validation set. The rows in Figure~\ref{fig:multi_future} show predicted futures for lane changes to left or keep the current lane (row 1), driving towards the roadside and straight ahead (row 2), and left/right turns at intersections (row 3).

\subsection{Generation of Diverse Multiview Videos}
\methodname{} can function as a multiview video generator conditioned on temporal layouts. 
This enables applications as a neural simulator for \methodname{}. 
Although trained on nuScenes~\cite{caesar2020nuscenes} \emph{train} set, \methodname{} exhibits creativity on the \emph{val} set by generating novel combinations of objects, motions, and scenes.

\begin{figure*}[ht]
    \includegraphics[width=\linewidth]{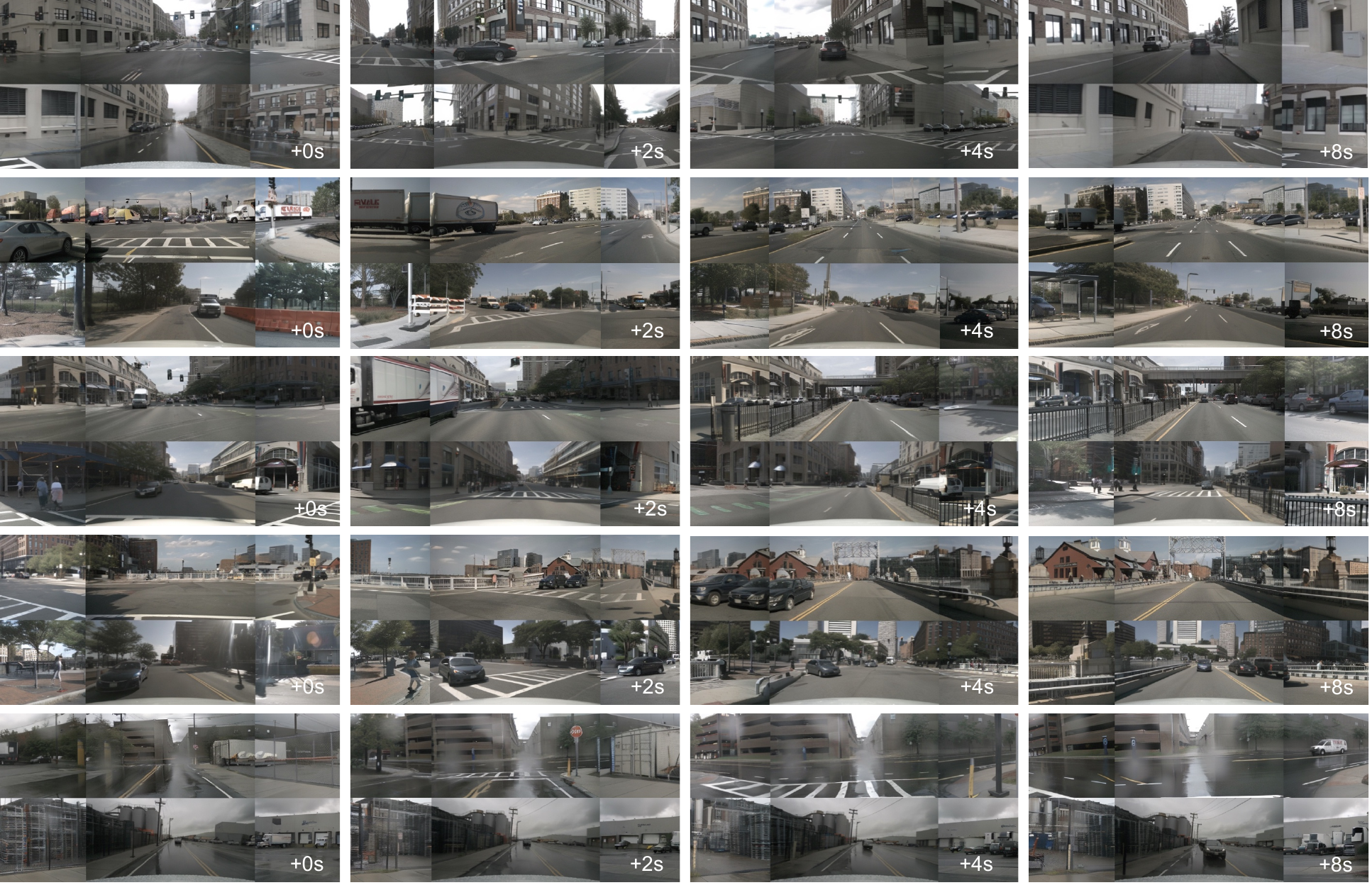}
    \caption{\textbf{Conditional generation of diverse multiview videos.} Given layout conditions (3D box, HD map, and BEV segmentation) from the nuScenes val set, our model is able to generate spatio-temporal consistent multiview videos.}
    \label{fig:id_video}
\end{figure*}

\paragraph{Normal scenes generation.}
\methodname{} can generate diverse multiview video forecasts based on layout conditions, as shown for the nuScenes~\cite{caesar2020nuscenes} validation set in Figure~\ref{fig:id_video}.

\begin{figure*}[ht]
    \includegraphics[width=\linewidth]{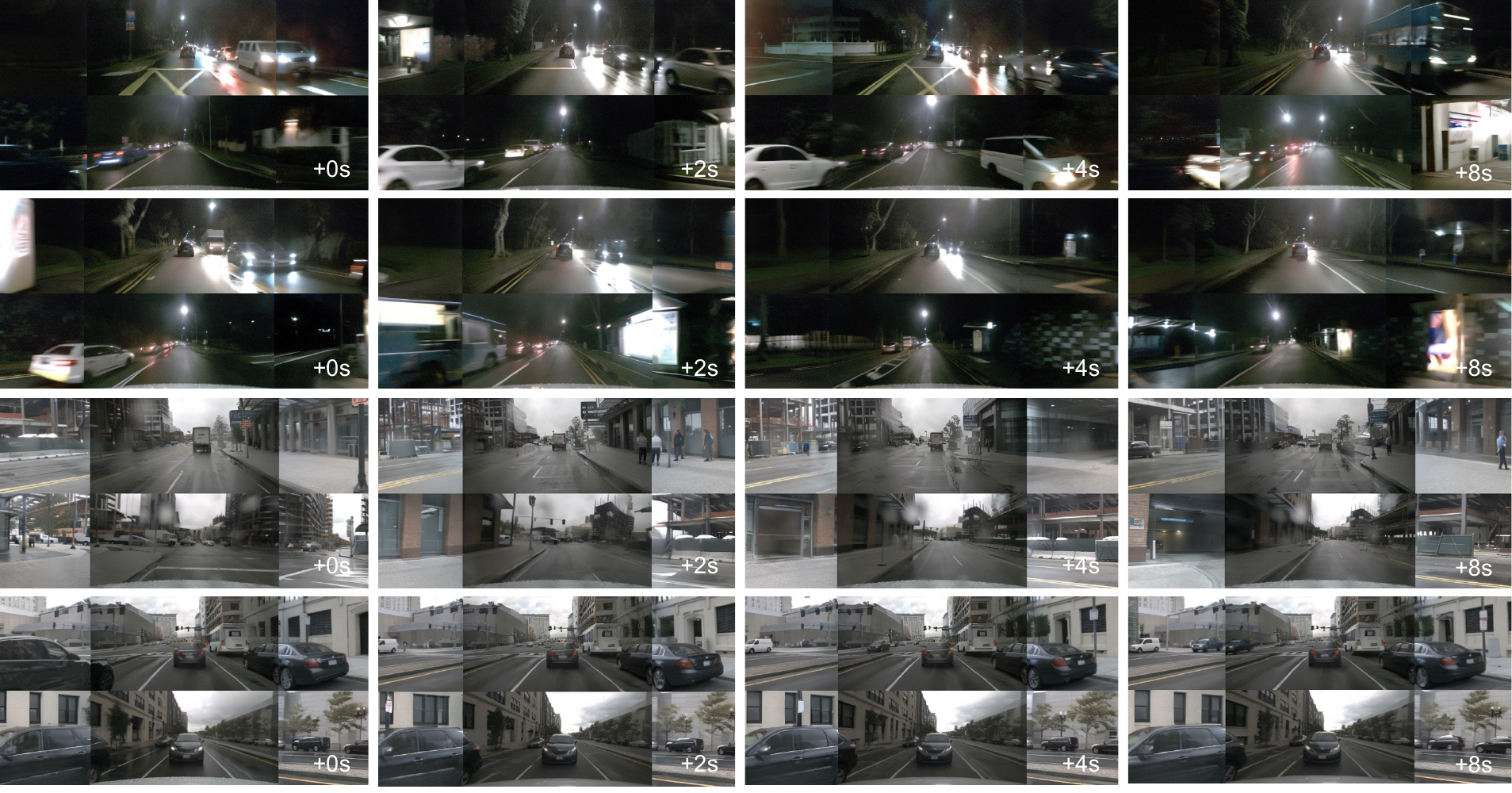}
    \caption{\textbf{Rare scenes generation.} Top two rows: night scenarios. Bottom two rows: rainy scenarios.}
    \label{fig:rare}
\end{figure*}

\paragraph{Rare scenes generation.}
It can also produce high-quality videos for rare driving conditions like nighttime and rain, despite limited exposure during training, as illustrated in Figure~\ref{fig:rare}. This demonstrates the model's ability to generalize effectively beyond the daytime scenarios dominant in the training data distribution.

\subsection{Visual Element Control}
\methodname{} allows conditional generation through various forms of control, including text prompts to modify global weather and lighting, ego-vehicle actions to change driving maneuvers, and 3D boxes to alter foreground layouts. This section demonstrates \methodname{}'s flexible control mechanisms for interactive video generation based on user-specified conditions. 

\begin{figure*}[ht]
    \includegraphics[width=\linewidth]{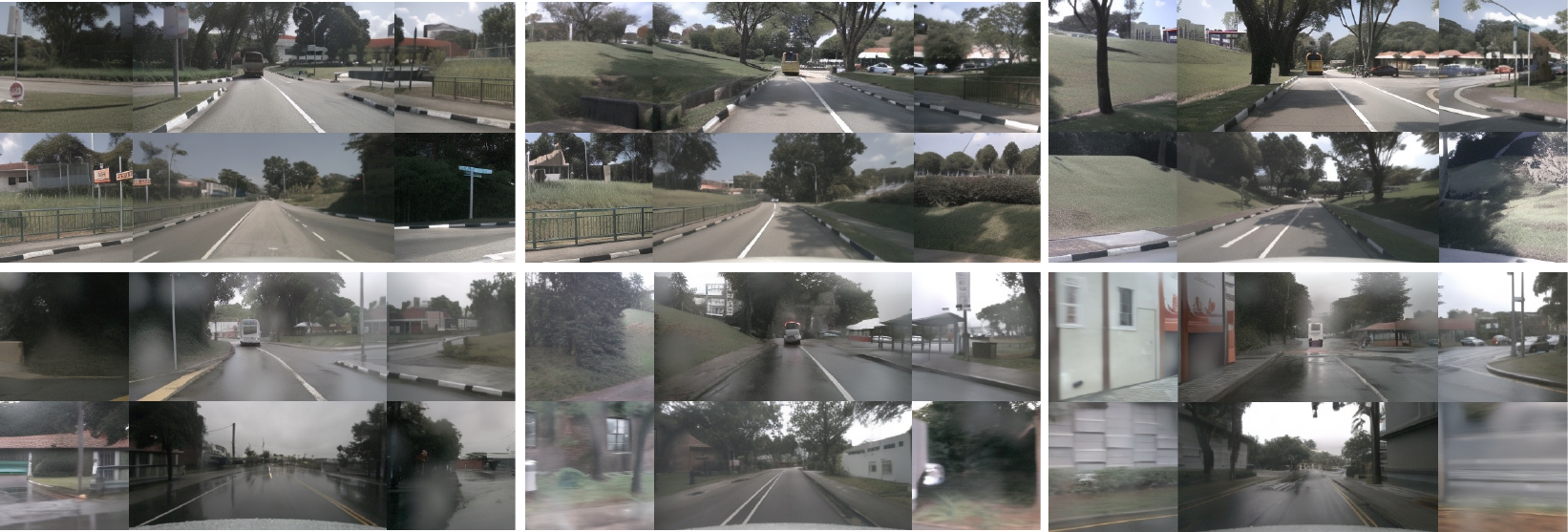}
    \caption{\textbf{Weather change generation.} The top row displays sunny daylight scenes. The bottom row shows the same layouts rendered as rainy scenes, demonstrating conditional generation capabilities.}
    \label{fig:weather}
\end{figure*}

\begin{figure*}[ht]
    \includegraphics[width=\linewidth]{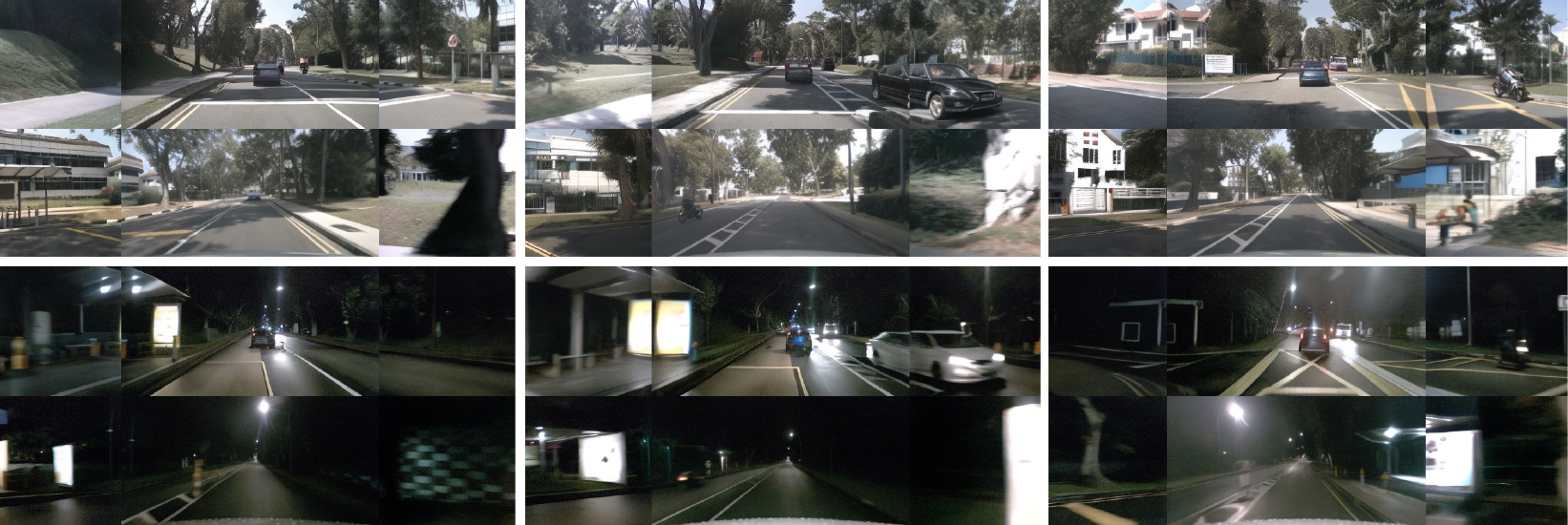}
    \caption{\textbf{Lighting change generation.} The top row displays daytime scenes. The bottom row shows the same layouts rendered as nighttime scenes, demonstrating conditional generation capabilities.}
    \label{fig:light}
\end{figure*}

\paragraph{Weather \& Lighting change.}
As shown in Figure~\ref{fig:weather} and Figure~\ref{fig:light}, we demonstrate the ability of our model to change weather or lighting conditions while maintaining the same scene layout (road structure and foreground objects).
Video examples are generated based on the layout conditions from nuScene \emph{val} set. This ability has great potential for future data augmentation. By generating diverse scenes under various weather and lighting conditions, our model can significantly expand the training dataset, thereby improving the generalization performance and robustness of the model.

\begin{figure*}[ht]
    \includegraphics[width=\linewidth]{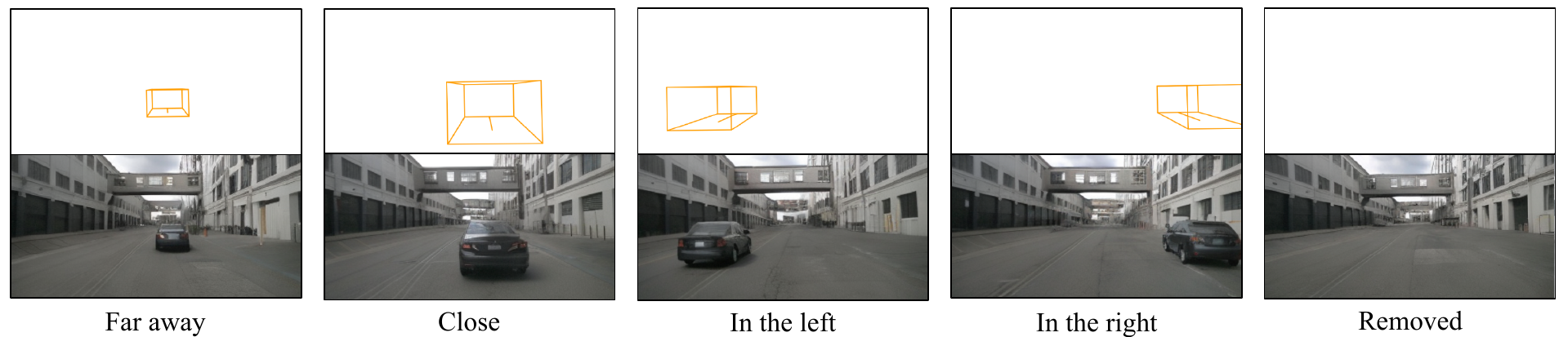}
    \caption{\textbf{Control foreground object layouts.} By modifying the positions of 3D boxes, high-fidelity images are produced that correspond to the layout changes specified.}
    \label{fig:control_obj}
\end{figure*}

\paragraph{Action control.}
\begin{figure*}
    \centering
    \includegraphics[width=\linewidth]{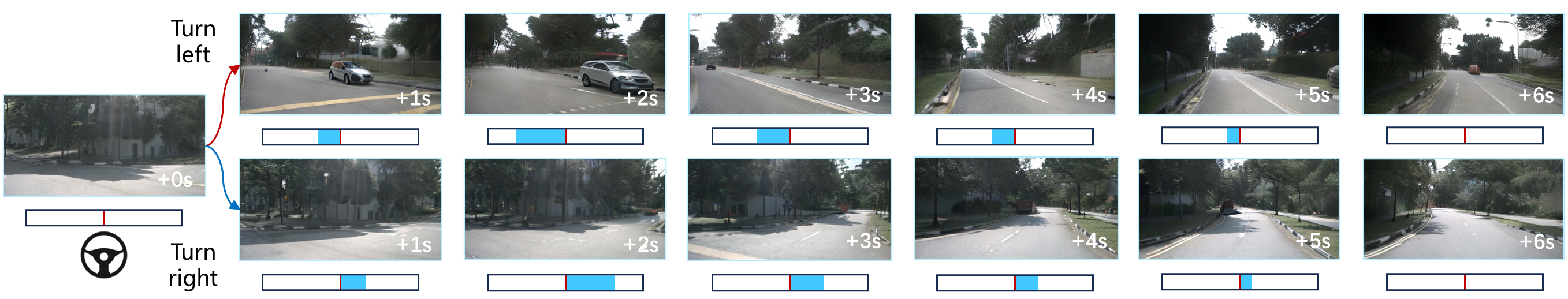}
    \caption{\textbf{Diverse turning behaviors.} Utilizing an identical initial frame, we provide our model with sequences of positive steering angles (indicating a left turn) and negative values (indicating a right turn). The figure demonstrates the model's proficiency in generating consistent street views for both turning behaviors. Each frame is accompanied by a blue bar, indicating the corresponding steering angle in degrees. A longer bar correlates with a more substantial steering angle. For clarity, only the front view is shown.}
    \label{fig:diverse_turning}
\end{figure*}

 \begin{figure*}
    \centering
    \includegraphics[width=\linewidth]{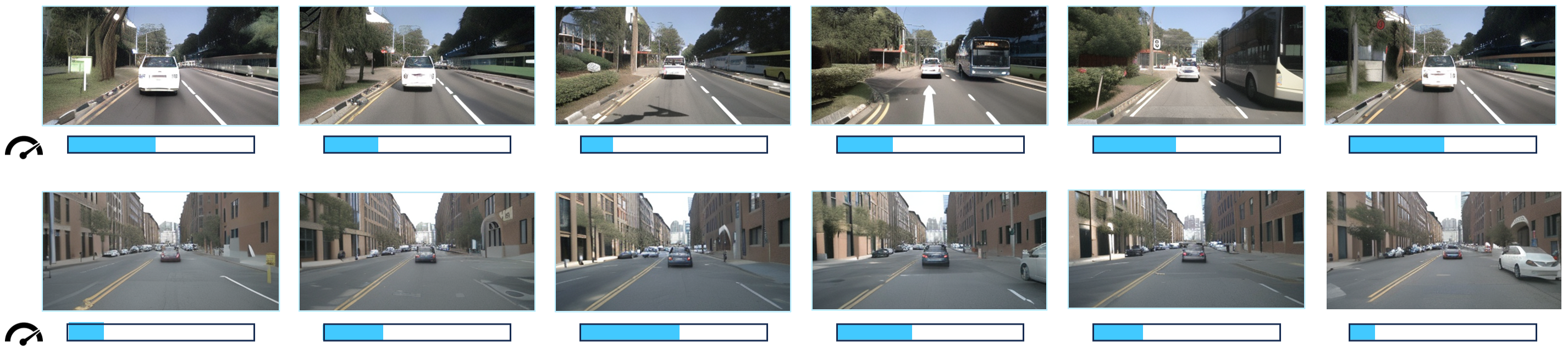}
    \caption{\textbf{Diverse speeding behaviors.} We input different patterns of speed signals into our model to assess controllability in terms of speed. The top series shows that the ego car decelerates and then accelerates while the bottom one shows a contrary behavior. These two results highlight the realism of our model's prediction.}
    \label{fig:diverse_speed}
\end{figure*}

Our model is capable of generating high-quality street views consistent with given ego-action signals. For instance, as shown in~\cref{fig:diverse_turning}, our model correctly generates turning-left and turning-right videos from the same initial frame according to the input steering signals. In~\cref{fig:diverse_speed}, our model successfully predicts the positions of the surrounding vehicles conforming with both the accelerating and decelerating signals. These qualitative results demonstrate the high controllability of our world model.

\paragraph{Foreground control.}
As Figure~\ref{fig:control_obj} shows, \methodname{} enables fine-grained control of foreground layouts in generated videos. By modifying lateral and longitudinal conditions, high-fidelity images are produced that correspond to the layout changes specified.

\begin{figure*}
    \includegraphics[width=\linewidth]{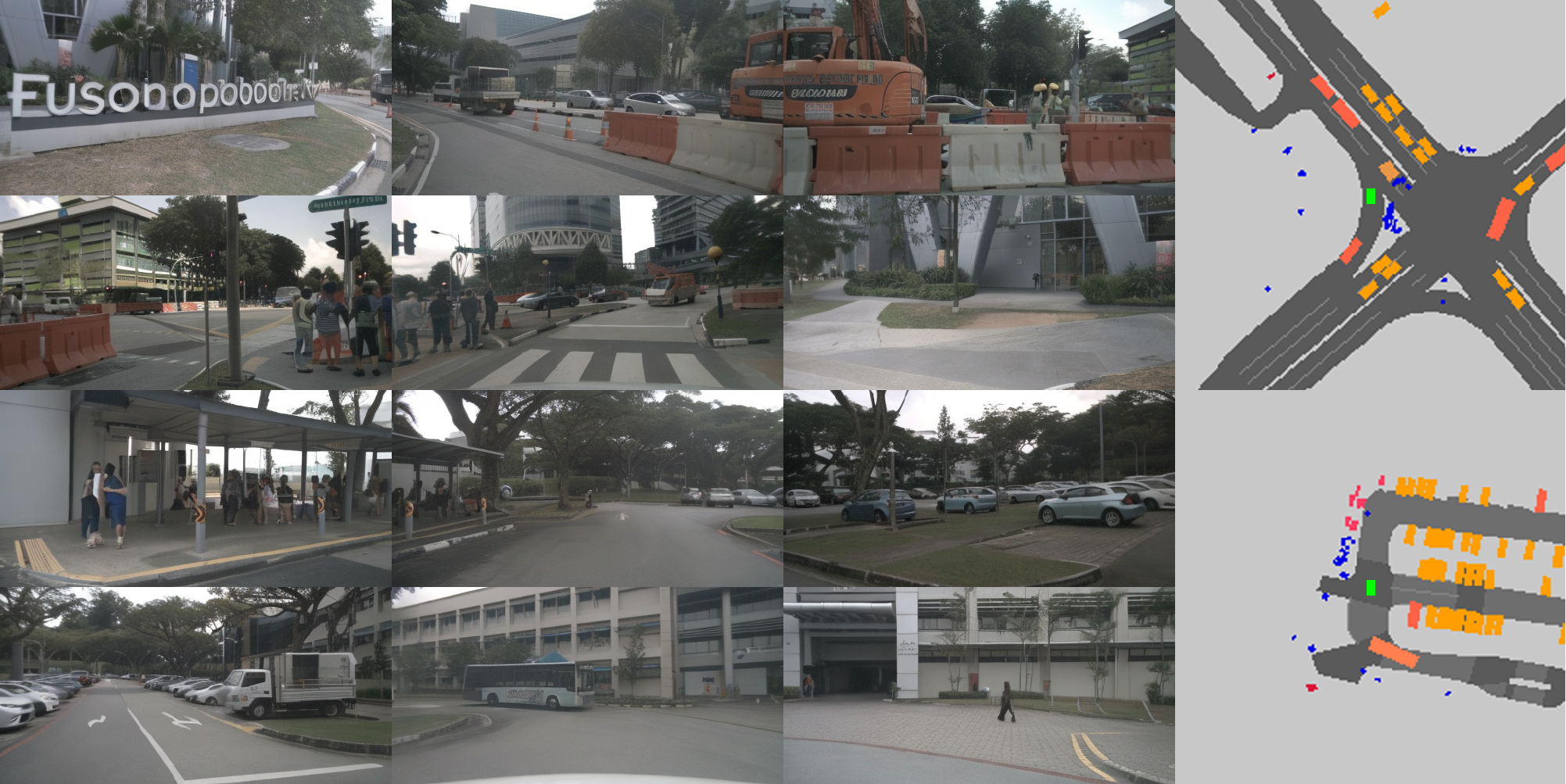}
    \caption{\textbf{Pedestrian generation examples.} }
    \label{fig:ped_multiview}
\end{figure*}

Pedestrian generation poses challenges for street-view synthesis methods. However, unlike previous work~\cite{swerdlow2023street, yang2023bevcontrol}, Figure~\ref{fig:ped_multiview} shows \methodname{} can effectively generate pedestrians. The first six images displays a vehicle waiting for pedestrians to cross, while the second six images shows pedestrians waiting at a bus stop. This demonstrates our model's potential to produce detailed multi-agent interactions.

\subsection{End-to-end Planning Results for Out-of-domain Scenarios}
\begin{figure*}[ht]
    \includegraphics[width=\linewidth]{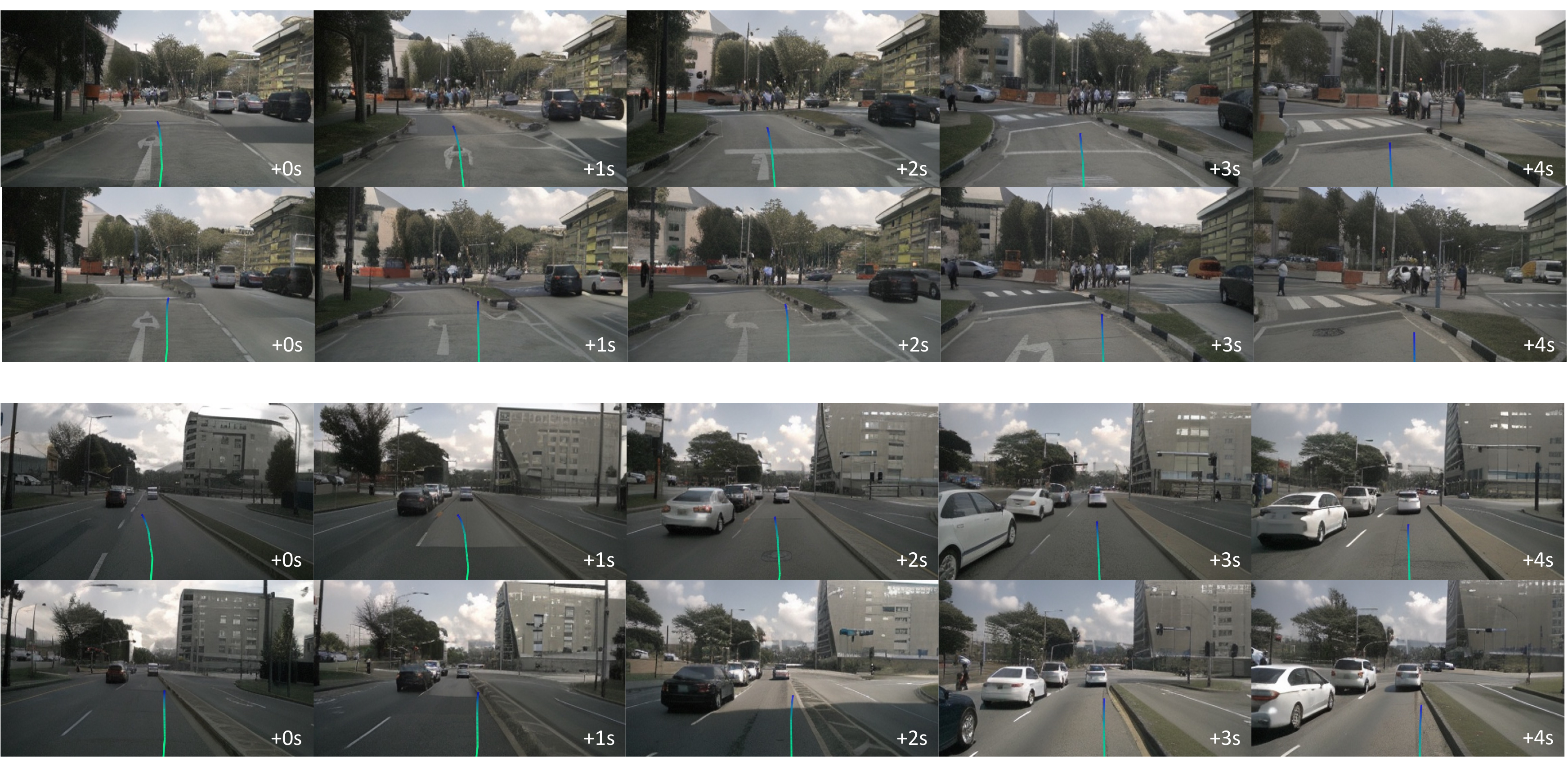}
    \caption{\textbf{Videos demonstrating the VAD planning results under normal and out-of-domain cases.}  We shift the ego location 0.5m to the right to create an out-of-domain case. The top row of each scene: the reasonable trajectory prediction of the VAD method under normal data. The bottom row of each scene: the irrational trajectory when encountering out-of-distribution cases.}
    \label{fig:ood_video}
\end{figure*}

Existing end-to-end planners are trained on expert trajectories aligned to lane centers. As Figure~\ref{fig:planner deviation} shows, this causes difficulties generating off-center deviations, known as the ``lack-of-exploration" problem in behavior cloning~\cite{codevilla2019exploring}. Using \methodname{} for simulation, Figure~\ref{fig:ood_video} demonstrates more planned trajectories deviating from the lane center. We find the planner from \cite{Jiang_2023_ICCV} cannot recover when evaluated on these generated out-of-distribution cases. This highlights the utility of \methodname{} for exploring corner cases and improving robustness.

\begin{figure*}[ht]
    \includegraphics[width=\linewidth]{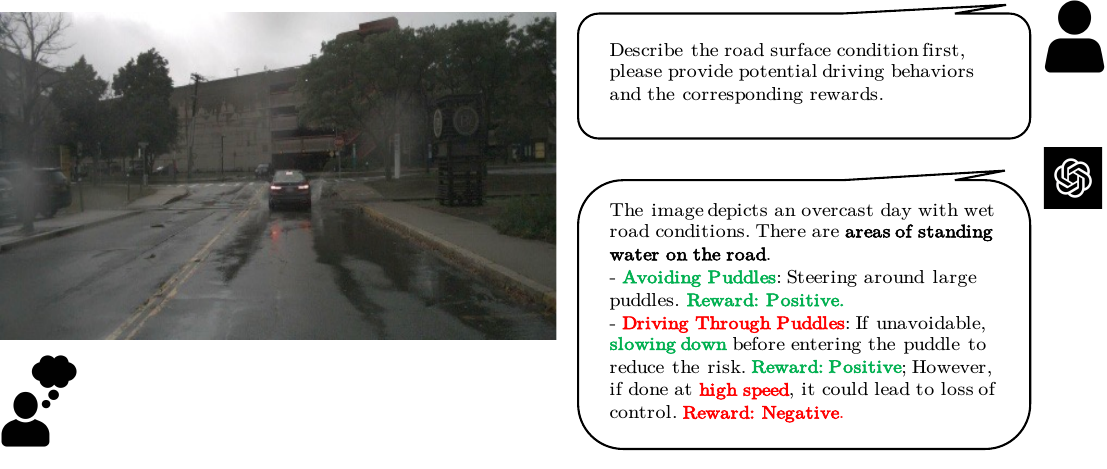}
    \caption{\textbf{GPT-4V as a reward function.} GPT-4V could give more reasonable rewards in cases such as there is a puddle ahead on the road.}
    \label{fig:gpt4v}
\end{figure*}

\subsection{Using GPT-4V as a Reward Function}
To assess the safety of different futures forecasted under different plans, we leverage the recent GPT-4V model as an evaluator. Specifically, we use \methodname{} to synthesize diverse future driving videos with varying road conditions and agent behaviors. We then employ GPT-4V to analyze these simulated videos and provide holistic rewards in terms of driving safety.
As illustrated in Figure~\ref{fig:gpt4v}, it demonstrates different driving behaviors that GPT-4V plans when there is a puddle ahead on the road.
Compared to reward functions with vectorized inputs, GPT-4V provides a more generalized understanding of hazardous situations in the \methodname{} videos. By deploying GPT-4V's multimodal reasoning capacity for future scenario assessment, we enable enhanced evaluation that identifies risks not directly represented but inferred through broader scene understanding.
This demonstrates the value of combining generative world models like \methodname{} with reward-generating models like GPT-4V. By using GPT-4V to critique \methodname{}'s forecasts, more robust feedback can be achieved to eventually improve autonomous driving safety under diverse real-world conditions.

\subsection{Video Generation on Other Datasets}

\begin{figure*}[ht]
    \includegraphics[width=\linewidth]{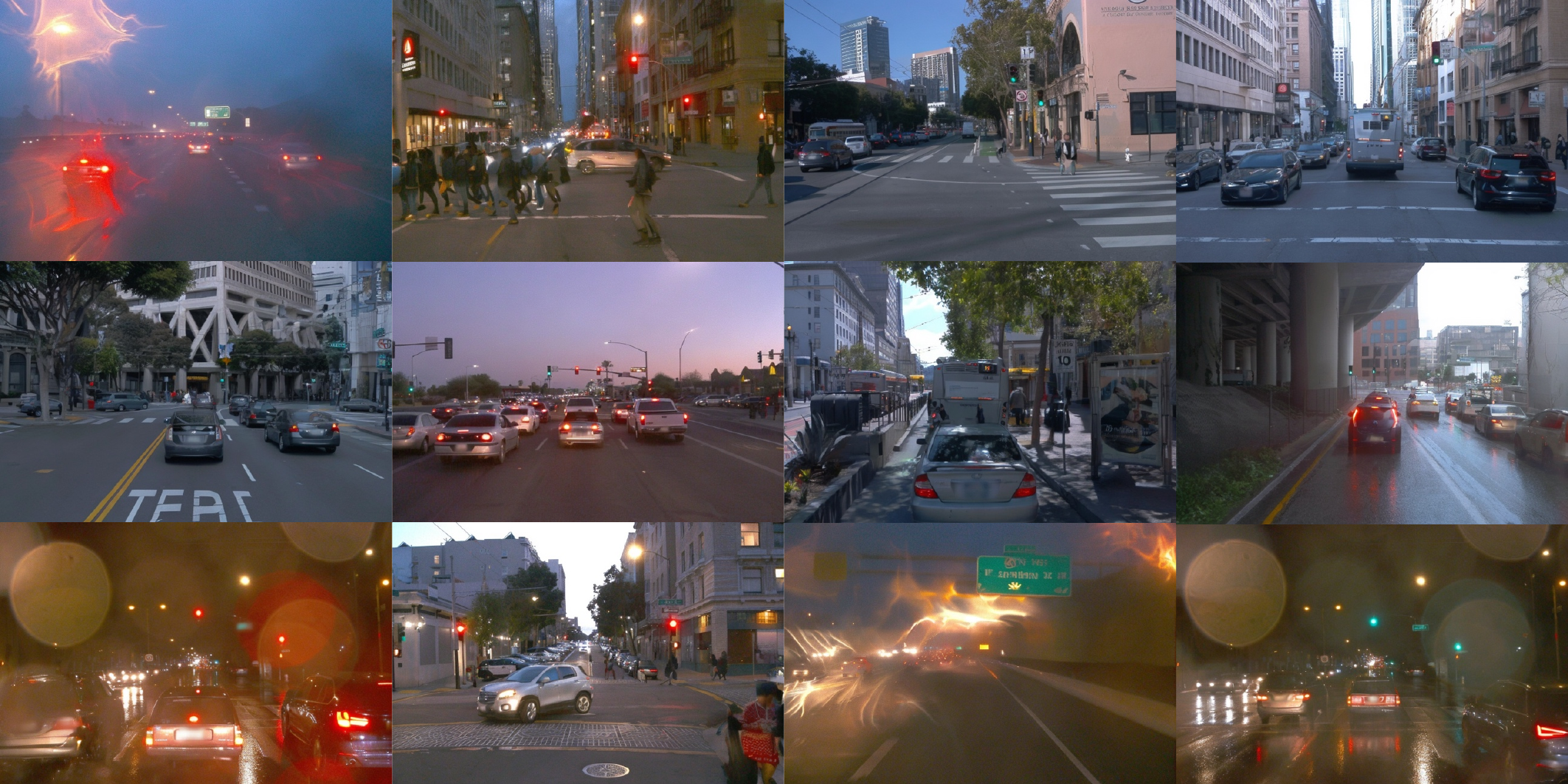}
    \caption{\textbf{Generation of High-Resolution Image on Waymo Open Dataset.} We showcase the image generation results for a wide range of traffic density, driving scenarios, lighting, and weather conditions.}
    \label{fig:waymo_hd_img}
    \vspace{-10pt}
\end{figure*}

\begin{figure*}[ht]
    \includegraphics[width=\linewidth]{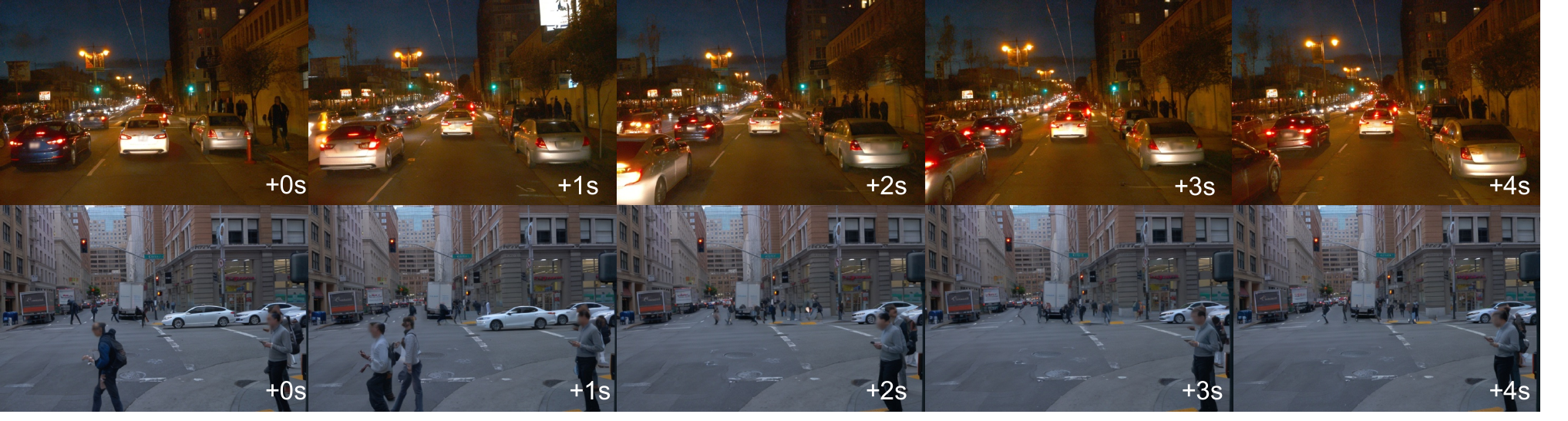}
    \caption{\textbf{Generation of High-Resolution Video on Waymo Open Dataset.} We showcase the video generation results in highly interactive driving scenarios like closely following the front car in heavy traffic (top) or waiting for pedestrians to cross the road.}
    \label{fig:waymo_hd_video}
\end{figure*}

\paragraph{Waymo Open Dataset.}
To showcase the wide applicability of \methodname{}, we apply it to generate high-resolution 768×512 images on the Waymo Open Dataset. As seen in Figure~\ref{fig:waymo_hd_img} and Figure~\ref{fig:waymo_hd_video}, \methodname{} produces realistic and diverse driving forecasts at this resolution with the same hyper-parameters for nuScenes. By generalizing effectively to new datasets and resolutions, these Waymo examples verify that \methodname{} provides a widely adaptable approach to high-fidelity video synthesis across different driving datasets. 

\section{Implementation Details}
In this section, we introduce the training \& inference details of 
\emph{joint multiview video model} and \emph{factorization model}.

\subsection{Joint Multiview Video Model}
\label{sec:details_joint}
\paragraph{Training Details.}
The original image resolution of nuScenes is 1600 $\times$ 900. We initially crop it to 1600 $\times$ 800 by discarding the top area and then resize it to 384 $\times$ 192 for model training.
Similar to VideoLDM~\cite{blattmann2023align}, we begin by training a conditional image latent diffusion model.
The model is conditioned on various scene elements, such as HD maps, BEV segmentation, 3D bounding boxes, and text descriptions.
All the conditions are concatenated in the token-length dimension. 
The image model is initialized with Stable Diffusion checkpoints~\cite{rombach2022high}
This \emph{conditional image model} is trained for 60,000 iterations with a total batch size of 768. We use the AdamW optimizer with a learning rate $1\times10^{-4}$.
Subsequently, we build the \emph{multiview video model} by introducing temporal and multiview parameters (Sec.~\ref{sec:joint_modeling}) and fine-tune this model for 40,000 iterations with a batch size of 32, with video frame length $T=8$. 
For action-based video generation, the difference lies only in the change of condition information for each frame, while the rest of the training and model structure are the same.
We use the AdamW optimizer~\cite{kingma2014adam} with a learning rate $5\times10^{-5}$ for the video model.
To sample from our models, we generally use the sampler from Denoising Diffusion Implicit Models (DDIM)~\cite{song2020denoising}. 
Classifier-free guidance (CFG) reinforces the impact of conditional guidance. For each condition, we randomly drop it with a probability of 20\% during training.
All experiments are conducted on A40 GPUs. 

\paragraph{Inference Details.}
During inference, the number of sampling steps is 50, and we use stochasticity $\eta$=1.0, CFG=5.0.
For video generation, we use the first frame as the condition to generate subsequent video content. Similar to VideoLDM~\cite{blattmann2023align}, we use the generated frame as the subsequent condition for long video generation.

\subsection{Factorization Model}
\paragraph{Training implementation of factorization.}
The implementation is overall similar to the implementation of the joint modeling in Sec.~\ref{sec:details_joint}. 
For the factorized generation, we additionally use reference views as extra image conditions.

Taking nuScenes data as an example, we first sort the six multiview video clips clockwise, denoted as $\x_0$ to $\x_5$.
Then a training sample is defined as $\{\x_{(i-1)\bmod6}, \x_i, \x_i^\prime, \x_{(i+1)\bmod6}\}$. where $\x_i$ is the stitched view randomly sampled from all six views, and $\{\x_{(i-1)\bmod6}, \x_{(i+1)\bmod6}\}$ is a pair of reference view. $\x_i^\prime$ is the previously generated frame of $i$-th view, which also serves as an additional image condition.
The training pipeline is very similar to the joint modeling, while here we generate a single view every iteration instead of multiple views.
As can be seen from the training sample $\{\x_{(i-1)\bmod6}, \x_i, \x_i^\prime, \x_{(i+1)\bmod6}\}$, we have view dimension $N = 1$ for the single stitch view $\x_i$ in training.
\vspace{-3mm}
\paragraph{Inference of factorization.}
During inference, we pre-define some views as the reference views, and the corresponding videos of these reference views are first generated by the joint model.
Then we generate the videos of stitched views conditioned on the paired reference video clips and previously generated view (i.e., $\x_i^\prime$).
Particularly, in nuScenes, we select $\texttt{F, BL, BR}$\footnote{F: front; B: back; L: left; R: right} as reference views.
The three reference views constitute three pairs, serving as the condition for three stitched views.
For example, our model generates front-left views conditioning on front views, back-left views, and previously generated front-left views.
The inference parameters are the same with Sec.~\ref{sec:details_joint}.

\section{Data}
In this section, we first describe the dataset preparation and then introduce the curation of the dataset to enhance the action-based generation.

\subsection{Data Preparation}
\paragraph{NuScenes Dataset.}
The nuScenes dataset provides full 360-degree camera coverage and is currently a primary dataset for 3D perception and planning. Following the official configuration, we use 700 street-view scenes for training and 150 for validation.
Next, we introduce the processing of each condition. For the 3D box condition, we project the 3D bounding box onto the image plane, utilizing octagonal corner points to depict the object's position and dimensions, while colors are employed to distinguish different categories. This ensures accurate object localization and discrimination between different objects.
For the HD map condition, we project the vector lines onto the image plane, with colors indicating various types.
In terms of the BEV segmentation, we adhere to the generation process outlined in CVT~\cite{zhou2022cross}. This process generates a bird's-eye view segmentation mask, which represents the distribution of different objects and scenery in the scene.
For the text condition, we sift through information provided in each scene description. 
For the planning condition, we utilize the ground truth movement of the ego locations for training, and the planned output from VAD~\cite{Jiang_2023_ICCV} for inference. This allows the model to learn from accurate ego-motion information and make predictions that are consistent with the planned trajectory.
Finally, for the ego-action condition, we extracted the information of vehicle speed and steering for each frame.

\paragraph{The Waymo Open Dataset.}
The Waymo Open Dataset~\cite{sun2020scalability} is a well-known large-scale dataset for autonomous driving. We only utilize data from the "front" camera to train the video model, with an image resolution of $768\times512$ pixels. For the map condition, we follow the data processing in OpenLane~\cite{chen2022persformer}.

\subsection{Data Curation}
\begin{figure*}
    \centering
    \includegraphics[width=\linewidth]{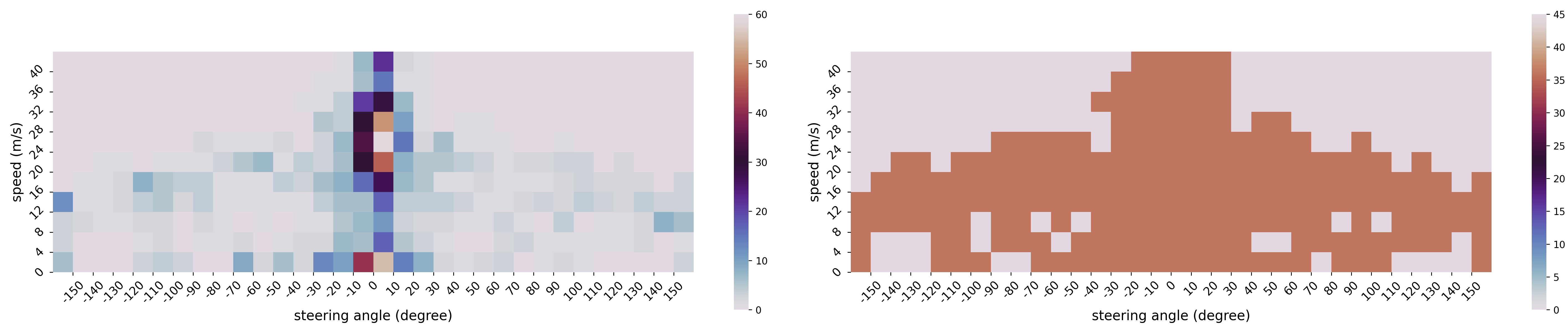}
    \caption{\textbf{The ego-action distribution before (left) and after re-sampling (right).} We re-sample rare combinations of speeds and steering angles, obtaining a balanced training dataset.}
    \label{fig:resampling}
\end{figure*}
 
The ego action distribution of the nuScenes dataset is heavily imbalanced: a large proportion of its frames exhibit small steering angles (less than 30 degrees) and a normal speed in the range of 10-20 m/s. This imbalance leads to weak generalizability to rare combinations of steering angles and speeds.

To alleviate this negative impact, we balance our training dataset by re-sampling rare ego actions. Firstly, we split each trajectory into several clips, each of which demonstrates only one type of driving behavior (i.e., turning left, going straight, or turning right). This process results in 1048 unique clips. Afterward, we cluster these clips by digitizing the combination of average steering angles and speeds. The speed range [0, 40] (m/s) is divided into 10 bins with equal lengths. Extreme speeds greater than 40 m/s will fall into the 11th bin. The steering angle range [-150, 150] (degree) is divided into 30 bins with equal lengths. Likewise, extreme angles greater than 150 degrees or less than -150 degrees will fall into another two bins, respectively. We plot the ego-action distribution resulting from this categorization in~\cref{fig:resampling}.

To balance the action distribution of these clips, we sample $N=36$ clips from each bin of the 2D $32\times11$ grid. For a bin containing more than $N$ clips, we randomly sample $N$ clips; For a bin containing fewer than $N$ clips, we loop through these clips until $N$ samples are collected. Consequently, 7272 clips are collected. The action distribution after re-sampling can be seen in~\cref{fig:resampling}. 

\section{Metric Evaluation Details}
\subsection{Video Quality}
The FID and FVD calculations are performed on 150 validation video clips from the nuScenes dataset. Since our model can generate multiview video, we break it down into six views of video for evaluation. We have a total of 900 video segments (~40 frames) and follow the calculation process described in VideoLDM~\cite{blattmann2023align}. We use the official UCF FVD evaluation code\footnote{\url{https://github.com/SongweiGe/TATS/}}.

\subsection{KPM Illustration}
\begin{figure*}[ht]
    \includegraphics[width=\linewidth]{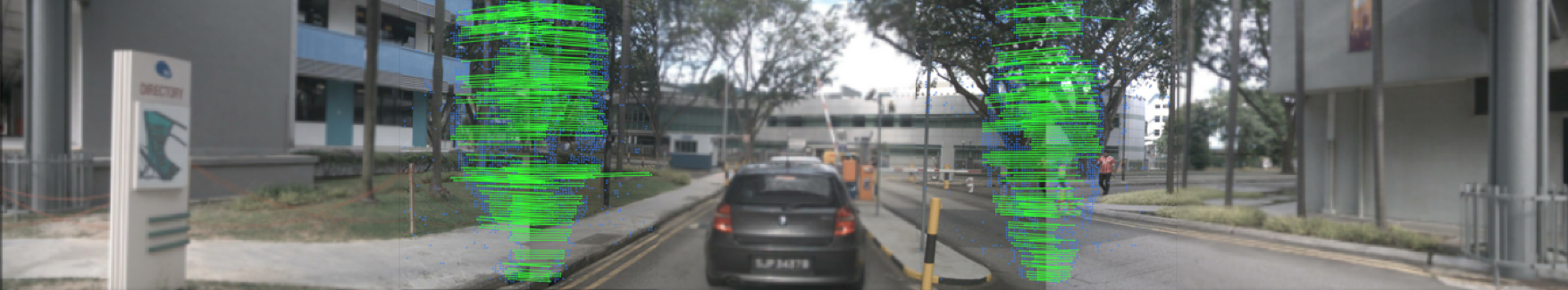}
    \caption{\textbf{Illustration of keypoint matching in KPM calculation.} The blue points are the image keypoints in the overlapping regions on the left/right side of the image. The green lines are the matched keypoints between the current view and its two adjacent views using the LoFTR~\cite{sun2021loftr} matching algorithm. }
    \label{fig:kpm}
\end{figure*}

As mentioned in Sec.~\ref{sec:kpm}, we introduced the KPM score metric for measuring multiview consistency for generated images, which are not considered in both FID and FVD metrics.
As shown in Figure~\ref{fig:kpm}, we demonstrate the keypoint matching process. The blue points are the keypoints in the overlapping regions on the left/right side of the image. The green lines are the matched points between the current view and its two adjacent views using the LoFTR~\cite{sun2021loftr} matching algorithm.

\end{document}